\pdfoutput=1

\documentclass[11pt]{article}

\usepackage{acl}

\usepackage{times}
\usepackage{latexsym}

\usepackage[T1]{fontenc}

\usepackage[utf8]{inputenc}

\usepackage{microtype}

\usepackage{inconsolata}

\usepackage{amsmath}
\usepackage{amsfonts}  
\usepackage{graphicx}  
\usepackage{boldline}  
\usepackage{multirow}  
\usepackage{floatrow}  
\usepackage{subcaption} 
\usepackage{hyperref} 

\newcommand\blfootnote[1]{%
  \begingroup
  \renewcommand\thefootnote{}\footnote{#1}%
  \addtocounter{footnote}{-1}%
  \endgroup
}

\title{3D Rotation and Translation for Hyperbolic Knowledge Graph Embedding}

\author{Yihua Zhu$^{1,2}$ \qquad  Hidetoshi Shimodaira$^{1,2}$ \\
  $^1$Kyoto University \qquad $^2$RIKEN\\
  \texttt{zhu.yihua.22h@st.kyoto-u.ac.jp, shimo@i.kyoto-u.ac.jp}
  }

\date{}

\begin{document}
\maketitle

\begin{abstract}
    The main objective of Knowledge Graph (KG) embeddings is to learn low-dimensional representations of entities and relations, enabling the prediction of missing facts. A significant challenge in achieving better KG embeddings lies in capturing relation patterns, including symmetry, antisymmetry, inversion, commutative composition, non-commutative composition, hierarchy, and multiplicity. This study introduces a novel model called 3H-TH (3D Rotation and Translation in Hyperbolic space) that captures these relation patterns simultaneously.
    In contrast, previous attempts have not achieved satisfactory performance across all the mentioned properties at the same time. The experimental results demonstrate that the new model outperforms existing state-of-the-art models in terms of accuracy, hierarchy property, and other relation patterns in low-dimensional space, meanwhile performing similarly in high-dimensional space.

\end{abstract}

\blfootnote{Our code is available at \url{https://github.com/YihuaZhu111/3H-TH}.}

\section{Introduction} \label{sec:intro}

\begin{figure}[t]
\centering
\begin{subfigure}[b]{1\textwidth}
        \includegraphics[width=\textwidth]{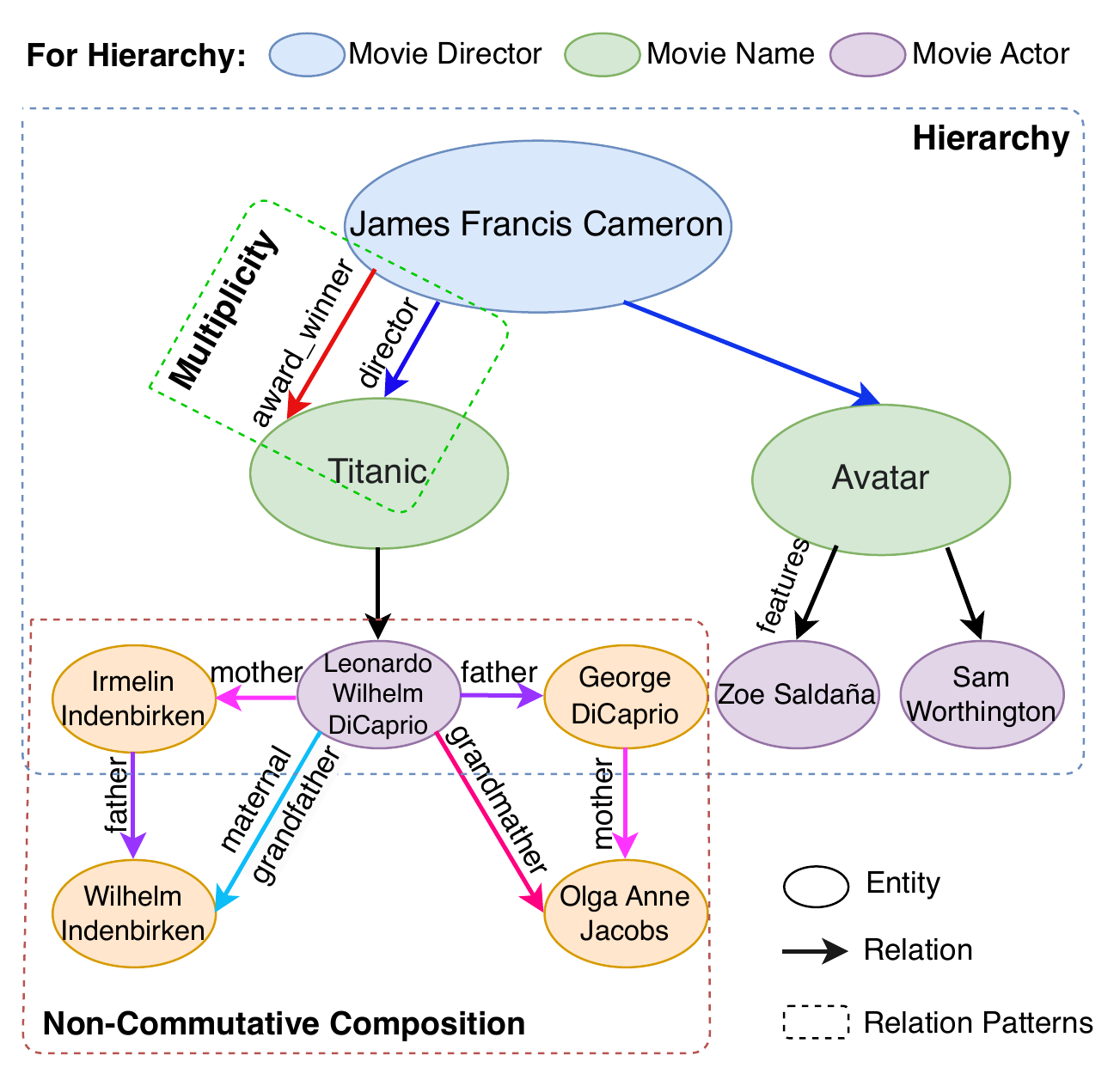}
    \end{subfigure}
\caption{Toy examples for three difficult relation patterns. Our approach can perform well in Hierarchy, Multiplicity, and Non-Commutative Composition.}
\label{fig:toy_example}
\end{figure}

The components of a knowledge graph are collections of factual triples, where each triple $(h, r, t)$ denotes a relation $r$ between a head entity $h$ and a tail entity $t$; toy examples are shown in Fig.~\ref{fig:toy_example}. 
Freebase \cite{bollacker2008freebase}, Yago \cite{suchanek2007yago}, and WordNet \cite{miller1995wordnet} are some examples of knowledge graphs used in the real world. Meanwhile, applications such as question-answering \cite{hao2017end}, information retrieval \cite{xiong2017explicit}, recommender systems \cite{zhang2016collaborative}, and natural language processing \cite{yang2019leveraging} may find significant value for knowledge graphs. Therefore, knowledge graph research is receiving increasing attention in both the academic and business domains.

Predicting missing links is a crucial aspect of knowledge graphs, given their typical incompleteness. In recent years, significant research efforts have focused on addressing this challenge through the utilization of knowledge graph embedding (KGE) techniques, which involve learning low-dimensional representations of entities and relations \cite{bordes2013translating,trouillon2016complex}. KGE approaches have demonstrated scalability and efficiency in modeling and inferring knowledge graph entities and relations based on available facts.

\begin{table*}[t]
\centering
\resizebox{\textwidth}{!}{\renewcommand{\arraystretch}{0.9}
\begin{tabular}{lccccccc}
\clineB{1-8}{2}
{Method} & {Symmetry} & {Antisymmetry}  & {Inversion} & {Commutative} & {Non-commutative} & {Hierarchy} & {Multiplicity}\\
\clineB{1-8}{2}
TransE (TE) &  & \checkmark & \checkmark & \checkmark &  &  &\\
RotatE (2E) & \checkmark & \checkmark & \checkmark & \checkmark &  &  &\\
QuatE (3E) & \checkmark & \checkmark & \checkmark & \checkmark & \checkmark &  &\\
MuRP (TH) &  & \checkmark & \checkmark & \checkmark & & \checkmark & \\
RotH (2H)& \checkmark & \checkmark & \checkmark & \checkmark &  & \checkmark & \checkmark \\
DualE & \checkmark & \checkmark & \checkmark & \checkmark & \checkmark &  & \checkmark \\
BiQUE & \checkmark & \checkmark & \checkmark & \checkmark &  & \checkmark &  \\
CompoundE & \checkmark & \checkmark & \checkmark & \checkmark &  &  & \checkmark \\
\textbf{(Proposal) 3H-TH} & \checkmark & \checkmark & \checkmark & \checkmark & \checkmark & \checkmark & \checkmark\\
\clineB{1-8}{2}
\end{tabular}
}
\caption{Relation patterns for existing and proposed models (\checkmark means ``can'')} 

\label{tab:relation_patterns}

\end{table*}

A major issue in KGE research concerned several relation patterns, including symmetry, antisymmetry, inversion, composition (i.e., commutative and non-commutative composition), hierarchy, and multiplicity (see Appendix \ref{subsec:relation_pattern_example}). In fact, several current approaches have attempted to model one or more of the above relation patterns \cite{bordes2013translating, sun2019rotate, chami2020low, cao2021dual}. The TransE \cite{bordes2013translating}, which models the antisymmetry, inversion, and composition patterns, represents relations as translations. The RotatE \cite{sun2019rotate} represents the relation as a rotation and aims to model symmetry, antisymmetry, inversion, and composition. For some difficult patterns (see Fig.~\ref{fig:toy_example}), including non-commutative composition, hierarchy, and multiplicity, the AttH \cite{chami2020low} embeds relation in hyperbolic space to enable relations to acquire hierarchy property. The DualE \cite{cao2021dual} attempts to combine translation and rotation operations to model multiple relations. Such approaches, however, have failed to perform well on all the above relation patterns simultaneously as shown in Table \ref{tab:relation_patterns}. Our proposed method 3H-TH, meaning 3D rotation in hyperbolic space and translation in hyperbolic space, can simultaneously model these relation patterns.

Here we present how our proposed method (3H-TH) works for the difficult relation pattern examples in Fig.~\ref{fig:toy_example}. By embedding the entities and relations in hyperbolic space, we can allow the KG model to acquire hierarchy properties so that we can more clearly distinguish between the different hierarchies of entities, for example, movie director, name, and actor. Besides, to solve non-commutative problems, for example (see Fig.~\ref{fig:toy_example}), if the mother of A's father (B) is C while the father of A's mother (D) is E, then C and E are equal if the relations were commutative, we use the quaternion geometry property (non-commutative) to enable the model to obtain a non-commutative composition pattern. Finally, we try to combine rotation and translation operations to obtain multiplicity properties, e.g. different relations exist between the same entities (e.g., \textit{award-winner, director}). 

Moreover, our study provides some important insights into developing several comparable methods to explore the impact of a combination of translation and rotation in Euclidean or hyperbolic space, as well as both simultaneously.
We evaluate the new model on three KGE datasets including WN18RR \cite{dettmers2018convolutional}, FB15K-237 \cite{toutanova2015observed}, and FB15K \cite{bordes2013translating}. Experimental results show that the new model outperforms existing state-of-the-art models in terms of accuracy, hierarchy property, and other relation patterns in low-dimensional space, meanwhile performing similarly in high-dimensional space, which indicates that the new model 3H-TH can simultaneously model symmetry, antisymmetry, inversion, composition, hierarchy, and multiplicity relation patterns.

\section{Related Work}
Knowledge graph embedding has received a lot of attention from researchers in recent years. One of the main KGE directions has been led by translation-based and rotation-based approaches. Another key area is hyperbolic KGE, which enables models to acquire hierarchy property. In particular, our approach advances in both directions and acquires both advantages.

\paragraph{Translation-based approach.} One of the widely adopted methods in KGE is the translation-based approach, exemplified by TransE \cite{bordes2013translating}, which represents relation vectors as translations in the vector space. In this approach, the relationship between the head and tail entities is approximated by adding the relation vector to the head entity vector, resulting in a representation that is expected to be close to the tail entity vector. After TransE, there has been an increasing amount of literature on its extension. TransH \cite{wang2014knowledge} represents a relation as a hyperplane to help the model perform better on complex relations. By embedding entities and relations in separate spaces with a shared projection matrix, TransR \cite{lin2015learning} further creates a relation-specific space to obtain a more expressive model for different types of entities and relations. Compared to TransR, TransD \cite{ji2015knowledge}  employs independent projection vectors for each object and relation, which can reduce the amount of computation. Although these methods are relatively simple and have only a few parameters, they do not effectively express crucial relation patterns such as symmetry, hierarchy, and multiplicity relations (Table \ref{tab:relation_patterns}).

\paragraph{Rotation-based approach.} RotatE \cite{sun2019rotate} introduced a new direction as rotation-based methods, which represents the relation vectors as rotation in complex vector space and can model various relation patterns, including symmetry, antisymmetry, inversion, and composition. QuatE \cite{zhang2019quaternion} substitutes 2D rotation with quaternion operation (3D rotation) in quaternion space, aiming to obtain a more expressive model than RotatE. Furthermore, the incorporation of 3D rotation enables the model to capture the non-commutative composition of relations, leveraging the geometric properties of quaternions (wherein two 3D rotations are known to be non-commutative). However, these rotation operations cannot solve hierarchy and multiplicity (Table \ref{tab:relation_patterns}).
DualE \cite{cao2021dual} presents a solution to the multiplicity problem by combining translation and rotation operations. However, the experimental results discussed in this paper do not provide conclusive evidence of the model's effectiveness in handling multiple relation data.
CompoundE \cite{ge-etal-2023-compounding} combines translation, 2D rotation, and scaling in Euclidean space to represent Knowledge Graphs, encompassing TransE \cite{bordes2013translating}, RotatE \cite{sun2019rotate}, LinearRE \cite{peng2020lineare}, and PairRE \cite{chao2020pairre} as its special cases. Although it captures various relation patterns, its limitation to 2D rotation and Euclidean space prevents it from capturing Non-commutative composition and Hierarchy properties.

\paragraph{Hyperbolic KGE.} One of the major challenges for KGE is the hierarchy problem. Hyperbolic geometry has been shown to provide an efficient approach to representing KG entities and relations in low-dimensional space while maintaining latent hierarchy properties. MuRP \cite{balazevic2019multi} optimizes the hyperbolic distance between the projected head entity and the translational tail entity to achieve comparable results by using fewer dimensions than the previous methods. RotH \cite{chami2020low} tries to substitute translation operations with rotation operations to obtain more relation patterns properties like RotatE. However, there is still room for improvement in handling other relation patterns, particularly in terms of multiplicity and non-commutative composition properties. BiQUE\cite{guo2021bique} utilizes biquaternions, which encompass both circular rotations in Euclidean space and hyperbolic rotations, aim to acquire hierarchy properties and RotatE-based relation patterns, while this approach struggles to effectively capture the Multiplicity property. Our proposed model 3H-TH leverages translation, 3D rotation, and hyperbolic embedding to offer a comprehensive and expressive representation of entities and relations, encompassing various relation patterns (Table \ref{tab:relation_patterns}).

\begin{table*}[t]
\centering
\resizebox{\textwidth}{!}{\renewcommand{\arraystretch}{1.1}
\begin{tabular}{lcccc}
\clineB{1-5}{2}
{Model} & {Relation embeddings} & {Translation} & {Rotation}  & {Scoring function}  \\
\clineB{1-5}{2}

        TransE (TE)   & $\mathbf{e}_{r}$ &E &         &  $-d^{E}\left( \mathbf{e}_{h} + \mathbf{e}_{r} , \mathbf{e}_{t} \right) + \! b_{h} \! + \!b_{t}$ \\
        RotatE (2E)   & $\mathbf{c}_{r}$ &  & 2D in E &  $-d^{E}\left( \mathbf{e}_{h} \circ \mathbf{c}_{r} , \mathbf{e}_{t} \right) + \! b_{h} \! + \!b_{t}$ \\
        QuatE (3E)    & $\mathbf{q}_{r}$ &  & 3D in E &  $(\mathbf{e}_{h} \otimes \mathbf{q}^{\triangleright}_{r} ) \cdot \mathbf{e}_{t} + \! b_{h} \! + \!b_{t}$\\
        MuRP (TH)     & $\mathbf{b}_{r}$ & H &         &  $-d^{\xi_{r}}\left(\mathbf{b}_{h} \oplus^{\xi_{r}} \mathbf{b}_{r}, \mathbf{b}_{t}\right)^{2} \! + \! b_{h} \! + \!b_{t}$\\
        RotH (2H)     & $\mathbf{c}_{r}$ &  & 2D in H &  $-d^{\xi_{r}}\left(\mathbf{b}_{h} \circ \mathbf{c}_{r}, \mathbf{b}_{t}\right)^{2} \! + \! b_{h} \! + \!b_{t}$ \\
        \textbf{3H}         & $\mathbf{q}_{r}$ &   & 3D in H &  $-d^{\xi_{r}}\left(\mathbf{b}_{h} \otimes \mathbf{q}_{r}, \mathbf{b}_{t}\right)^{2} \! + \! b_{h} \! + \!b_{t}$ \\
\hline
        2E-TE & $\mathbf{c}_{r}, \mathbf{e}_{r}$ & E & 2D in E &  $-d^{E}\left( \mathbf{e}_{h} \circ \mathbf{c}_{r} + \mathbf{e}_{r}, \mathbf{e}_{t} \right) + \! b_{h} \! + \!b_{t}$ \\ 
        3E-TE & $\mathbf{q}_{r}, \mathbf{e}_{r}$ & E & 3D in E &  $-d^{E}\left( \mathbf{e}_{h} \otimes \mathbf{q}^{\triangleright}_{r} + \mathbf{e}_{r}, \mathbf{e}_{t} \right) + \! b_{h} \! + \!b_{t}$\\  
        2E-TE-2H-TH & $\mathbf{c}_{(r,E)}, \mathbf{e}_{r}, \mathbf{c}_{(r,H)}, \mathbf{b}_{r}$ & E, H & 2D in E, H & $-d^{\xi_{r}} \left( \left( \mathbf{b}_{\gamma} \circ \mathbf{c}_{(r,H)} \right) \oplus^{\xi_{r}} \mathbf{b}_{r} , \mathbf{b}_{t} \right)^{2} \! + \! b_{h} \! + \!b_{t}$   \\ 
        \textbf{3H-TH} & $\mathbf{q}_{r}, \mathbf{b}_{r}$ & H & 3D in H &  $-d^{\xi_{r}} \left( \left( \mathbf{b}_{h} \otimes \mathbf{q}^{\triangleright}_{r} \right) \oplus^{\xi_{r}} \mathbf{b}_{r} , \mathbf{b}_{t} \right)^{2} \! + \! b_{h} \! + \!b_{t}$ \\ 
        3E-TE-3H-TH & $\mathbf{q}_{(r,E)}, \mathbf{e}_{r}, \mathbf{q}_{(r,H)}, \mathbf{b}_{r}$ & E, H & 3D in E, H & $-d^{\xi_{r}} \left(  \left( \mathbf{b}_{\lambda} \otimes \mathbf{q}^{\triangleright}_{(r,H)} \right) \oplus^{\xi_{r}} \mathbf{b}_{r} , \mathbf{b}_{t} \right)^{2} \! + \! b_{h} \! + \!b_{t}$ \\  
\clineB{1-5}{2}
\end{tabular} 
}
\caption{Six component models and examples of composite models. 
3H is a new component model for 3D rotation in hyperbolic space.
The composite model 3H-TH performed best in the experiment.
E and H in the table represent Euclidean and hyperbolic space, respectively.
$\mathbf{q}^{\triangleright}_{r}$ denotes normalization, $\circ$ denotes Hadamard product, and $\otimes$ denotes Hamilton product.
Also, 
$\mathbf{b}_{\gamma} := \mathbf{e}_{h} \circ \mathbf{c}_{(r,E)} + \mathbf{e}_{r}$ and
$\mathbf{b}_{\lambda} := \mathbf{e}_{h} \otimes \mathbf{q}^{\triangleright}_{(r,E)} + \mathbf{e}_{r}$ are used to simplify the formula.
}
\label{tab: comparable_model}
\end{table*}


\section{Problem Formulation and Background}
We describe the KGE problem and present some related methods before our approach part.

\subsection{Knowledge graph embedding} \label{sec:kge}
Given a knowledge graph with a set of fact triples $(h, r, t) \in \mathcal{E} \subseteq \mathcal{V} \times \mathcal{R} \times \mathcal{V}$, where $\mathcal{V}$ and $\mathcal{R}$ represent sets of entities and relations, respectively. Mapping entities $v\in\mathcal{V}$ to embeddings $\mathbf{e}_v$ in $k_\mathcal{V}$ dimensions and relations $r\in\mathcal{R}$ to embeddings $\mathbf{e}_r$ in $k_\mathcal{R}$ dimensions is the goal of KGE.

We use the scoring function $s:\mathcal{V}\times \mathcal{R}\times\mathcal{V}\rightarrow\mathbb{R}$ to measure the difference between the transformed entities and target entities, and the difference is mainly composed of distance including Euclidean distance: 
\[
d^{E}\left( \mathbf{x}, \mathbf{y} \right) = \left\| \mathbf{x}-\mathbf{y} \right\|
\]
and hyperbolic distance \cite{ganea2018hyperbolic}:
\begin{align} \label{eq:distance_hyperbolic}
   d^{\xi_{r}}\left( \mathbf{x}, \mathbf{y} \right) = \frac{2}{\sqrt{\xi_{r}}}\mathrm{tanh}^{-1}(\sqrt{\xi_{r}}||-\mathbf{x}\oplus^{\xi_{r}}\mathbf{y}||), 
\end{align}
where $\left\| \cdot \right\|$, $\oplus^{\xi_{r}}$, and $\xi_{r}$ represent L2 norm, M\"obius addition (see Equation \ref{eq:mobius_addition}), and curvature in hyperbolic space, respectively.

\subsection{TransE}
Inspired by word2vec \cite{mikolov2013efficient} in the domain of word embedding, TransE \cite{bordes2013translating} is the first translation-based work in the field of KGE, representing relations as translations in Euclidean space. Given triple vectors ($ \mathbf{e}_{h} \in \mathbb{R}^{k}, \mathbf{e}_{r} \in \mathbb{R}^{k}, \mathbf{e}_{t} \in \mathbb{R}^{k} $), the scoring function of TransE is
\[
s = -d^{E}\left( \mathbf{e}_{h} + \mathbf{e}_{r} , \mathbf{e}_{t} \right),
\]
then maximize $s$ to train this model.

\subsection{2D and 3D rotation}
To enable KGE models to acquire more relation patterns, including symmetry, antisymmetry, inversion, and composition, RotatE \cite{sun2019rotate} represents relation as 2D rotation in complex space $\mathbb{C}$. Given triple vectors ($ \mathbf{e}_{h} \in \mathbb{R}^{k}, \mathbf{c}_{r} \in \mathbb{C}^{\frac{k}{2}}, \mathbf{e}_{t} \in \mathbb{R}^{k} $), the scoring function of RotatE is
\[s = -d^{E}\left( \mathbf{e}_{h} \circ \mathbf{c}_{r} , \mathbf{e}_{t} \right),\] where the elements of $\mathbf{c}_r$ are constrained to be on the unit circle in $\mathbb{C}$, i.e., $ \left| (\mathbf{c}_{r})_i \right| = 1 $, and the symbol $ \circ $ denotes Hadamard product. 

QuatE \cite{zhang2019quaternion} replaces 2D rotation with a quaternion operation (3D rotation) in quaternion space $\mathbb{Q}$, with the aim of obtaining a more expressive model than RotatE. Given $ \mathbf{e}_{h} \in \mathbb{R}^{k}, \mathbf{q}_{r} \in \mathbb{Q}^{\frac{k}{4}}, \mathbf{e}_{t} \in \mathbb{R}^{k} $, the scoring function of QuatE is 
\[
s = (\mathbf{e}_{h} \otimes \mathbf{q}^{\triangleright}_{r} ) \cdot \mathbf{e}_{t}
\]
Where  $ \mathbf{q}^{\triangleright}_{r} $, $ \otimes $, and $ \cdot $ represent quaternion normalization, Hamilton product, and dot product, respectively (see Appendix \ref{subsec:Hamilton’s_quaternions}).

\subsection{Hyperbolic geometry}

We give a brief summary of hyperbolic geometry, and all the hyperbolic geometry equations that we need to use are shown in Appendix \ref{subsec:Hyperbolic_Geometry}, including the logarithmic transformation $\mathrm{log}_\mathbf{0}^{\xi_{r}}(\mathbf{v})$, the exponential transformation $\mathrm{exp}_\mathbf{0}^{\xi_{r}}(\mathbf{y})$, and the M\"obius addition ($x \oplus^{\xi_{r}} y$).

MuRP \cite{balazevic2019multi} is the first paper to introduce translation in hyperbolic space $\mathbb{B}$. Given triple vectors ($ \mathbf{b}_{h} \in \mathbb{B}^{k}, \mathbf{b}_{r} \in \mathbb{B}^{k}, \mathbf{b}_{t} \in \mathbb{B}^{k} $), the scoring function is
\[
s = -d^{\xi_{r}}\left(\mathbf{b}_{h} \oplus^{\xi_{r}} \mathbf{b}_{r}, \mathbf{b}_{t}\right)^{2} \!,
\]
where $\oplus^{\xi_{r}}$ and $d^{\xi_{r}}(.,. )$ represent M\"obius addition and hyperbolic distance respectively.

RotH \cite{chami2020low} aims to replace translation operations with rotation operations in hyperbolic space, similar to how RotatE operates in Euclidean space, in order to capture additional relational patterns. Given triple vectors ($ \mathbf{b}_{h} \in \mathbb{B}^{k} $, $ \mathbf{c}_{r} \in \mathbb{C}^{\frac{k}{2}} $, $ \mathbf{b}_{t} \in \mathbb{B}^{k} $), the scoring function is defined as
\[
s = -d^{\xi_{r}}\left(\mathbf{b}_{h} \circ \mathbf{c}_{r}, \mathbf{b}_{t}\right)^{2} \!,
\]
where the elements of $\mathbf{c}_{r}$ are constrained to be on the unit circle in $\mathbb{C}$.

\section{Our Approach} \label{sec:our_approach}

Our proposed model aims to enhance the representation of entities and relations by incorporating various relation patterns, with a particular focus on non-commutative composition, multiplicity, and hierarchy. To achieve this, we leverage techniques such as translation, 3D rotation, and hyperbolic embedding, allowing for a more expressive and comprehensive representation.

\subsection{Component models}

To maintain a concise representation of the component models for translation and rotation, we have adopted a straightforward naming convention using two letters. The first letter indicates the type of operation: T for translation, 2 for 2D rotation, and 3 for 3D rotation. The second letter indicates the space: E for Euclidean space and H for hyperbolic space. For example, TE represents translation (T) in Euclidean space (E). In total, there are $3\times 2 = 6$ possible combinations of component models that serve as building blocks for creating composite models. The pipeline of any composite model is created by concatenating the component models. Further details regarding various component models and composite models can be found in Table~\ref{tab: comparable_model}.

In the preceding sections, we have introduced TransE (TE), RotatE (2E), QuatE (3E), MuRP (TH), and RotH (2H). Another model not yet proposed is 3H, which does 3D rotation in hyperbolic space. In this study, we propose a new rotation model 3H as follows.
Given triple vectors $ \mathbf{b}_{h} \in \mathbb{B}^{k}, \mathbf{q}_{r} \in \mathbb{Q}^{\frac{k}{4}}, \mathbf{b}_{t} \in \mathbb{B}^{k}$, the scoring function of 3H is
\[
s = -d^{\xi_{r}}\left(\mathbf{b}_{h} \otimes \mathbf{q}^{\triangleright}_{r}, \mathbf{b}_{t}\right)^{2} \!.
\]

\subsection{3H-TH model} \label{subsec:3H-TH}

When examining Table~\ref{tab:relation_patterns}, we can observe that 3D rotation is essential for capturing non-commutative properties, while hyperbolic space is crucial for representing hierarchy. Additionally, combining 2d rotation and translation plays an important role in capturing multiplicity; we can expect that the new extension of 3H-TH (3D rotation and translation) possesses similar properties. Taking all these factors into consideration, we will investigate the 3H-TH model that combines these essential elements.

Given head entity $ \mathbf{e}_{h} \in \mathbb{R}^{k} $ and tail entity  $ \mathbf{e}_{t} \in \mathbb{R}^{k} $, as well as the relation that is split into a 3D rotation part $ \mathbf{q}_{r} \in \mathbb{Q}^{\frac{k}{4}} $ and a translation part $ \mathbf{e}_{r} \in \mathbb{R}^{k} $, we map entities $\mathbf{e}_{h}, \mathbf{e}_{t}$ and the translation relation $ \mathbf{e}_{r} $ from Euclidean space ($ \mathbf{e}_{h}, \mathbf{e}_{t}, \mathbf{e}_{r} \in \mathbb{R}^{k}$) to hyperbolic space ($  \mathbf{b}_{h}, \mathbf{b}_{t}, \mathbf{b}_{r} \in \mathbb{B}^{k}$) using the exponential transformation:
\begin{align} \label{eq:3hth-b}
    \mathbf{b}_{\delta} = \mathrm{exp}_\mathbf{0}^{\xi_{r}}(\mathbf{e}_{\delta}) \in \mathbb{B}^{k}, \delta = h,r,t.
\end{align}
as detailed in Equation \ref{eq:expmap}.

The utilization of hyperbolic space in KG models enables the acquisition of hierarchical properties. It is important to note that each relation $r$ in the KG has a unique curvature $\xi_{r}$ \cite{chami2020low}. Unlike MuRP, where all relations have the same curvature, we train different values of curvature $\xi_{r}$ for relation $r$ to represent varying degrees of curvature in the hyperbolic space. A higher value of $\xi_{r}$ for a specific relation signifies a greater degree of hierarchy, resembling a tree-like structure. Conversely, a flatter space represents less hierarchy in the corresponding relation.

The non-commutative property of 3D rotation enables the KG model to perform non-commutative composition, making it more expressive compared to 2D rotation. Therefore, we apply the 3D rotation operation (\textbf{3H}) to the mapped head entity in hyperbolic space. Additionally, using rotation and translation operations alone does not allow the model to acquire the multiplicity property. However, combining rotation and translation enables the KG model to exhibit multiplicity. Thus, we utilize M\"obius addition ($x \oplus^{\xi_{r}} y$) as Euclidean translation in hyperbolic space (\textbf{TH}). The final operation of 3H-TH model is represented as follows:
\begin{align}
  \mathbf{b}_{ \left( \mathbf{e}_{h}, \mathbf{e}_{r}, \mathbf{q}_{r} \right) } = \left( \mathbf{b}_{h} \otimes \mathbf{q}^{\triangleright}_{r} \right) \oplus^{\xi_{r}} \mathbf{b}_{r}. \label{eq:3H-TH}
\end{align}
Here, $ \otimes $ and $ \mathbf{q}^{\triangleright}_{r} $ represent the Hamilton product and normalization, respectively.

\subsection{Scoring function and loss} \label{subsec:scoring_function}

We utilize the hyperbolic distance between the final transformed head entity $ \mathbf{b}_{ \left( \mathbf{e}_{h}, \mathbf{e}_{r}, \mathbf{q}_{r} \right) } $ and the mapped tail entity $ \mathbf{b}_{t} $ as the scoring function:
\begin{align}
   s(h,r,t) = -d^{\xi_{r}} \left( \mathbf{b}_{ \left( \mathbf{e}_{h}, \mathbf{e}_{r}, \mathbf{q}_{r} \right) } , \mathbf{b}_{t} \right)^{2} \! + \! b_{h} \! + \!b_{t}.
\end{align}
Here, $d^{\xi_{r}}(.)$ is the hyperbolic distance introduced in Equation~\ref{eq:distance_hyperbolic} with the curvature $\xi_{r}$, and $b_{v}(v \in \mathcal{V})$ represents the entity bias added as a margin in the scoring function \cite{tifrea2018poincar, balazevic2019multi}. The comparison of various scoring functions, encompassing hyperbolic distance-based, Euclidean distance-based, and dot product-based methods, is detailed in Appendix~\ref{subsec:scoring-function-discussion}.
Moreover, instead of using other negative sampling methods, we uniformly select negative instances for a given triple $(h, r, t)$ by perturbing the tail entity. The model is trained by minimizing the full cross-entropy loss, defined as follows:
\begin{align}
L =\sum_{t^{\prime}} \log \left(1+\exp \left(y_{t^{\prime}} \cdot s\left(h, r, t^{\prime}\right)\right)\right) \label{eq:loss-function} \\
y_{t^{\prime}}=\left\{\begin{array}{l}
-1, \text { if } t^{\prime}=t  \nonumber \\
1, \text { otherwise }
\end{array}\right.
\end{align}

\subsection{Other composite models}

We have introduced a novel component model called 3H, which involves 3D rotation in hyperbolic space. We have also developed a composite model called 3H-TH, which combines 3D rotation and translation in hyperbolic space, as discussed earlier. Furthermore, we have created several other composite models (as shown in Table \ref{tab: comparable_model}), including 2E-TE (2D Rotation and Translation in Euclidean space), 3E-TE (3D Rotation and Translation in Euclidean space), 2E-TE-2H-TH (2D Rotation and Translation in both Euclidean and Hyperbolic space), and 3E-TE-3H-TH (3D Rotation and Translation in both Euclidean and Hyperbolic space).

To examine the effects of integrating translation and rotation, we compare 2E-TE and 3E-TE with their respective counterparts, 2E and 3E. Additionally, we compare 2E-TE-2H-TH and 3E-TE-3H-TH with RotH and 3H-TH to investigate the effects of operations in different spaces. These comparisons allow us to analyze the contributions and implications of different components in the models.

We provide a detailed explanation of 3E-TE-3H-TH because the other models are interpreted as a part of this most complex model. 
Embeddings of head and tail entities are $ \mathbf{e}_{h},\mathbf{e}_{t} \in \mathbb{R}^{k} $, and embeddings of relation $r$ are $  \mathbf{q}_{(r,E)} \in \mathbb{Q}^{\frac{k}{4}}, \mathbf{e}_{(r, E)}\in \mathbb{R}^{k}, \mathbf{q}_{(r,H)}\in \mathbb{Q}^{\frac{k}{4}}, \mathbf{e}_{(r, H)} \in \mathbb{R}^{k} $, where $ \mathbf{e}_{(r, \alpha)} $ and $ \mathbf{q}_{(r,\alpha)} $ are translation and 3D rotation relations, respectively, for space $\alpha \in \{E, H\}$.

We first perform 3D rotation and translation on the head entity in Euclidean space (\textbf{3E-TE}) using the following transformation:
\begin{align}
     \mathbf{e}_{ \left( \mathbf{e}_{h}, \mathbf{e}_{(r,E)}, \mathbf{q}_{(r,E)} \right) } = \left( \mathbf{e}_{h} \otimes \mathbf{q}^{\triangleright}_{(r,E)} \right) + \mathbf{e}_{(r,E)} 
\end{align}
Then we apply the same process as for 3H-TH (Equation \ref{eq:3H-TH}) to $ \mathbf{e}_{ \left( \mathbf{e}_{h}, \mathbf{e}_{(r,E)}, \mathbf{q}_{(r,E)} \right) }$, and we use the hyperbolic distance as the scoring function
\begin{align}
\begin{split}
   &s(h,r,t) = \\
   &-d^{\xi_{r}} \left(  \left( \mathbf{b}_{\lambda} \otimes \mathbf{q}^{\triangleright}_{(r,H)} \right) \oplus^{\xi_{r}} \mathbf{b}_{r} , \mathbf{b}_{t} \right)^{2} \! + \! b_{h} \! + \!b_{t}. \label{eq:scoring-3E-TE-3H-TH}
\end{split}
\end{align}
Finally, the loss function is defined by Equation~\ref{eq:loss-function} in Section~\ref{subsec:scoring_function}. We provide more details on several composite models in Table~\ref{tab: comparable_model}.

\section{Experiments}
We expect that the composite model 3H-TH, which performs both 3D rotation and translation in hyperbolic space, can effectively capture all relation patterns. We aim to validate this expectation through experimentation.

\begin{table}[t]
\resizebox{\textwidth}{!}{\renewcommand{\arraystretch}{1}
   \centering
   \begin{tabular}{lccccc}
   \clineB{1-6}{2}
   {Dataset} & {Entities} & {Relations} & {Train} & {Validation} & {Test} \\
   \clineB{1-6}{2}
   WN18RR & 40,943 & 11 & 86,835 & 3,034 & 3,134\\
   FB15k-237 & 14,541 & 237 & 272,115 & 17,535 & 20,466 \\
   FB15K & 14,951 & 1,345 & 483,142 & 50,000 & 59,071 \\
   \clineB{1-6}{2}
   \end{tabular}
   }
\caption{Details of the three datasets.}
\label{tab:datasets}
\end{table}

\begin{table*}[t]
\resizebox{\textwidth}{!}{\renewcommand{\arraystretch}{1}
    \centering
    \begin{tabular}{lllccccccccccc}
    \clineB{1-13}{2}
    & & \multicolumn{2}{c}{WN18RR} & \multicolumn{6}{c}{FB15k-237} & \multicolumn{2}{c}{FB15K} \\
     Model & MRR & H@1 & H@3 & H@10 & MRR & H@1 & H@3 & H@10 & MRR & H@1 & H@3 & H@10\\
     \clineB{1-13}{2}
        TransE(TE)  & .244 & .099 & .350 & .506 & .277 & .194 & .303 & .444 & .463 & .336 & .538 & .697 \\ 
        RotatE(2E)  & .387 & .330 & .417 & .491 & .290 & .208 & .316 & .458 & .469 & .355 & .527 & .691 \\ 
        QuatE(3E)  & .445 & .407 & .463 & .515 & .266 & .186 & .290 & .426 & .484 & .360 & .556 & .715 \\
        MuRP(TH)   & .269 & .106 & .402 & .532 & .279 & .196 & .306 & .445 & .486 & .358 & .565 & .718\\
        RotH(2H)  & .466 & .422 & .484 & .548 & .312 & .222 & .343 & .493 & .498 & .373 & .577 & .728 \\ 
        BiQUE  &   .298 & .231 & .328 & .425 & .309 & .223 & .339 & .479 & - & - & - & - \\
        3H     &  .467 & \underline{.429} & .486 & .541 & .277 & .195 & .302 & .444 & .500 & .375 & .576 & .726 \\
        2E-TE  & .448 & .421 & .474 & .522 & .262 & .184 & .283 & .419 & .494 & .373 & .568 & .725  \\ 
        3E-TE  & .456 & .408 & .467 & .518 & .261 & .184 & .282 & .414 & .496 & .376 & .572 & .725  \\ 
        2E-TE-2H-TH & \underline{.469} & .428 & \underline{.487} & \underline{\textbf{.552}} & .315 & .225 & \underline{.347} & .497 & .494 & .370 & .572 & .722  \\ 
        \textbf{3H-TH} & \textbf{.473} & \textbf{.432} & \textbf{.490} & \textbf{.552} & \textbf{.320} & \textbf{.229} & \textbf{.351} & \textbf{.501} & \textbf{.506} & \textbf{.383} & \textbf{.581} & \underline{.731}  \\ 
        3E-TE-3H-TH & .469 & .424 & .481 & .546 & \underline{.316} & \underline{.227} & .346 & \underline{.499} & \underline{.504} & \underline{.379} & \underline{.580} & \textbf{.733} \\ 
    \clineB{1-13}{2}
    \end{tabular}
    }
\caption{Link prediction accuracy results of three datasets in low-dimensional space ($k=32$). The best score is highlighted in \textbf{bold}, and the second-best score is \underline{underlined}. The \textbf{3H-TH} model outperforms other state-of-the-art methods significantly on WN18RR, FB15K-237, and FB15K. Results are statistically significant under paired student's t-test with p-value 0.05 except 2E-TE-2H-TH; more details refer to Appendix~\ref{subsec:t-test} }
    \label{tab:overview_result_in_low}
\end{table*}

\subsection{Experimental setup}
\paragraph{Dataset.} 
We evaluate our proposed method on three KG datasets, including WN18RR \cite{dettmers2018convolutional}, FB15K-237 \cite{toutanova2015observed}, and FB15K \cite{bordes2013translating} with licence CC-BY 2.5. The details of these datasets are shown in Table \ref{tab:datasets}. WN18RR is a subset of WN18 \cite{dettmers2018convolutional} which is contained in WordNet \cite{miller1995wordnet}. FB15K is a subset of Freebase \cite{bollacker2008freebase}, a comprehensive KG including data about common knowledge and FB15K-237 is a subset of FB15K. All three datasets were designed for KGE, and we employ them for KGE tasks, and all three datasets have no individual people or offensive content.

\paragraph{Evaluation metrics.} 
Given a head entity and a relation, we predict the tail entity and rank the correct tail entity against all candidate entities. We use two popular ranking-based metrics: (1) mean reciprocal rank (MRR), which measures the average inverse rank for correct entities: $\frac{1}{n} \sum_{i=1}^{n} \frac{1}{\text { Rank }_{i}}$. (2) hits on $K$ ($H \mathrm{@} K, K \in\{1,3,10\}$), which measures the proportion of correct entities appeared in the top $K$ entities.

\paragraph{Baselines.} We compare our new model with state-of-the-art (SOTA) methods, namely TransE \cite{bordes2013translating}, RotatE \cite{sun2019rotate}, QuatE \cite{zhang2019quaternion}, MuRP \cite{balazevic2019multi}, RotH \cite{chami2020low}, and BiQUE\cite{guo2021bique}. Alongside these five models and 3H-TH, our comparative models include 3H, 3E-TE, 2E-TE-3H-TH, and 3E-TE-3H-TH. It is worth noting that these comparative models have all been newly developed by us. Significantly, while hyperbolic-based methods indeed require longer training times compared to their Euclidean-based counterparts, it's worth noting that the space and time complexities of all these models remain equivalent. More details of state of the art baselines and discussion refer to Appendix~\ref{subsec:SoTA_KGE}.

\paragraph{Implementation.} The key hyperparameters in our implementation include the learning rate, optimizer, negative sample size, and batch size. To determine the optimal hyperparameters, we performed a grid search using the validation data. The optimizer options we considered are Adam \cite{kingma2014adam} and Adagrad \cite{duchi2011adaptive}. Finally, we obtain results by selecting the maximum values from three random seeds.

\begin{table*}[t]
\centering
   \resizebox{\textwidth}{!}{\renewcommand{\arraystretch}{1}
   \begin{tabular}{lccccccccc}
   \clineB{1-10}{2}
    &  \multicolumn{2}{c}{hierarchy measure} &  \multicolumn{6}{c}{} \\
   \cline{2-3}
   {Relation} & {$\text{Khs}_r$} & {$-\xi_r$} & {TE} & {2E} & {2H} & BiQUE & {2E-TE-2H-TH} & {\textbf{3H-TH}} & {3E-TE-3H-TH}\\
   \clineB{1-10}{2}
   member meronym & 1 & -2.9 & .407 & .304 & .390& .245 & \underline{.407} & \textbf{.412} & .391  \\
   hypernym & 1 & -2.46 & .192 & .235 & \underline{.251} & .164 & \textbf{.271} & .247 & .249  \\ 
   has part & 1 & -1.43 & .311 & .256 & \underline{.323} & .215 & .317 & .291 & \textbf{.337}  \\ 
   instance hypernym & 1 & -0.82 & .492 & .488 & .488 & \textbf{.529} & .488 & \underline{.503} & .500  \\ 
   member of domain region & 1 & -0.78 & .442 & .442 & \underline{.462} & .423 & .423 & \textbf{.465} & .423  \\ 
   member of domain usage & 1 & -0.74 & .417 & .438 & .438 & \textbf{.500} & .438 & \underline{.441} & .417  \\ 
   synset domain topic of & 0.99 & -0.69 & .428 & .399 & \underline{.430} & .386 & \textbf{.434} & .411 & .425  \\ 
   \hline
   also see & 0.36 & -2.09 & .732 & .625 & .652 & .598 & .652 & .637 & .634  \\ 
   derivationally related form & 0.07 & -3.84 & .959 & .960 & .961 & .784 & .966 & .960 & .960  \\ 
   similar to & 0.07 & -1 & 1 & 1 & 1 & .667 & 1 & 1 & 1  \\ 
   verb group & 0.07 & -0.5 & .962 & .974 & .974 & .654 & .974 & .974 & .962 \\ 
   \clineB{1-10}{2}
   \end{tabular}
   }
\caption{Link prediction accuracy results for specific relations sorted by $\text{Khs}_r$. Higher $\text{Khs}_r$ or lower $-\xi_r$ indicates a greater degree of hierarchy \cite{krackhardt2014graph}. Accuracy is measured by H@10 in low-dimensional space ($k=32$) for all 11 relations in WN18RR. The best score is highlighted in \textbf{bold}, and the second-best score is \underline{underlined}. We can observe that the 3H-TH model tends to perform well on relations with larger $\text{Khs}_r$ values, indicating its ability to capture hierarchical patterns.}
   \label{tab:low_hierarchy}
\end{table*}

\begin{table*}[t]
\resizebox{0.75\textwidth}{!}{\renewcommand{\arraystretch}{1}
    \centering
    \begin{tabular}{lccccc}
    \clineB{1-6}{2}
     Model & Symmetry & Antisymmetry & Composition & Inversion & multiplicity\\
     \clineB{1-6}{2}
        TransE(TE) & .321 & .335 & .362 & .511  & .643 \\ 
        RotatE(2E) & \textbf{.454} & \textbf{.497} & .338 & .512  & .663 \\ 
        QuatE(3E)  & .324 & .388 & .357 & .541 & .683  \\ 
        MuRP(TH)   & .335 & .359 & .361 & .542 & .666  \\
        RotH(2H)   & .360 & .441 & .366 & .558 & .686  \\ 
        3H & .357 & .458 & .363 & .559 & .685 \\
        2E-TE & .362 & \underline{.466} & .365 & .552 & .681  \\ 
        3E-TE & .361 & .465 & .366 & .557 & .689  \\ 
        2E-TE-2H-TH & .365 & .440 & .361 & .552 & .687  \\ 
        \textbf{3H-TH}  & \underline{.386} & .450 & \underline{.369} & \textbf{.566} & \textbf{.704} \\ 
        3E-TE-3H-TH & .361 & .444 & \textbf{.377} & \underline{.564} & \underline{.691} \\ 
    \clineB{1-6}{2}
    \end{tabular}
    }
\caption{Link prediction accuracy for specific relation patterns. Accuracy is measured by MRR for FB15K in low-dimensional space ($k=32$). \textbf{Bold} indicates the best score, and \underline{underline} represents the second-best score. The 3H-TH model achieves the best or second-best performance on the \textit{symmetry, composition, inversion,} and \textit{multiplicity} properties.}
    \label{tab:relation_patterns_accuracy}
\end{table*}

Moreover, to ensure a fair comparison, we incorporated entity bias ($b_{v}, v \in \mathcal{V}$) into the scoring function for all models (see Table \ref{tab: comparable_model}). Additionally, we used uniform negative sampling across all models. We give more details of implementation in Appendix~\ref{subsec:implementation}

Finally, we conduct additional experiments to examine the outcomes when we establish equal total parameters (see Appendix \ref{subsec:additional_composite_experiment}).

\subsection{Results in low dimensions}

Table \ref{tab:overview_result_in_low} provides an overview of the overall accuracy in low-dimensional space ($k=32$). Tables \ref{tab:low_hierarchy} and \ref{tab:relation_patterns_accuracy} present detailed results on hierarchy and relation patterns, respectively.

\paragraph{Overall accuracy.}
Table \ref{tab:overview_result_in_low} provides the link prediction accuracy results of WN18RR, FB15K-237, and FB15K in low-dimensional space ($k = 32$). The 3H-TH model outperforms all state-of-the-art models, particularly on the largest dataset FB15K, showcasing the powerful representation capacity achieved by combining 3D rotation and translation in hyperbolic space. Additionally, compared to RotH(2H), the 3H-TH model achieves competitive results across all evaluation metrics, indicating that 3D rotation in hyperbolic space enhances the model's expressiveness. Moreover, the 3H-TH model improves upon previous state-of-the-art Euclidean methods (RotatE and QuatE) by 6.1\%, 10.3\%, and 10.2\% in MRR on WN18RR, FB15K-237, and FB15K, respectively. This comparison highlights the superiority of hyperbolic geometry over Euclidean geometry in low-dimensional KG representation.

\begin{table*}[t]
\resizebox{\textwidth}{!}{\renewcommand{\arraystretch}{1}
    \centering
    \begin{tabular}{lllccccccccccc}
    \clineB{1-13}{2}
    & & \multicolumn{2}{c}{Dim = 200} & \multicolumn{6}{c}{Dim = 300} & \multicolumn{2}{c}{Dim = 500} \\
     Model & MRR & H@1 & H@3 & H@10 & MRR & H@1 & H@3 & H@10 & MRR & H@1 & H@3 & H@10\\
     \clineB{1-13}{2}
        TransE(TE)  & .263 & .107 & .380 & .532 & .262 & .108 & .379 & .531 & .260 & .104 & .378 & .532 \\ 
        RotatE(2E)  & .396 & .384 & .399 & .419 & .387 & .377 & .390 & .406 & .380 & .372 & .383 & .395 \\ 
        QuatE(3E)   & .487 & .442 & .503 & .573 & .490 & .444 & .506 & .580 & .490 & \underline{.443} & .507 & \textbf{.580}\\
        MuRP(TH)    & .265 & .105 & .392 & .531 & .263 & .102 & .388 & .529 & .260 & .102 & .380 & .529 \\
        RotH(2H)    & \underline{.490} & .444 & .507 & .578 & .488 & .443 & .506 & .575 & .489 & \underline{.443} & .508 & \underline{.579} \\ 
        3H          & .484 & .440 & .500 & .571 & \underline{.491} & \textbf{.447} & .507 & .576 & .487 & .441 & .503 & .575  \\
        2E-TE       & .393 & .382 & .396 & .415 & .390 & .379 & .395 & .411 & .383 & .372 & .388 & .400  \\ 
        3E-TE       & \underline{.490} & .445 & .506 & .578 & \textbf{.492} & .444 & \textbf{.511} & \underline{.581} & \textbf{.492} & \textbf{.445} & \underline{.509} & .585  \\ 
        2E-TE-2H-TH & \textbf{.493} & .446 & \underline{.509} & \underline{.585} & .490 & \underline{.446} & .505 & .578 & .489 & .442 & .507 & \underline{.579}  \\ 
        \textbf{3H-TH} & \textbf{.493} & \underline{.447} & \underline{.509} & \textbf{.587} & \underline{.491} & .443 & \textbf{.511} & \underline{.581} & \underline{.491} & \textbf{.445} & \textbf{.510} & \textbf{.580}  \\ 
        3E-TE-3H-TH & \textbf{.493} & \textbf{.448} & \textbf{.510} & .579 & \textbf{.492} & \underline{.446} & \underline{.508} & \textbf{.582} & .487 & \underline{.443} & .502 & .578  \\ 
    \clineB{1-13}{2}
    \end{tabular}
    }
\caption{The link prediction accuracy results of WN18RR in high-dimensional space ($k=200, 300, 500$). \textbf{Bold} indicates the best score, and \underline{underline} represents the second-best score.}
    \label{tab:overview_result_in_high_WN18RR}
\end{table*}

\begin{table*}[t]
    \resizebox{\textwidth}{!}{\renewcommand{\arraystretch}{0.9}
    \centering
    \begin{tabular}{lcccccccc}
    \clineB{1-9}{2}
    &  \multicolumn{2}{c}{hierarchy measure} &  \multicolumn{5}{c}{} \\
    \cline{2-3}
    {Relation} & {$\text{Khs}_r$} & {$-\xi_r$} & {TE} & {2E} & {2H} & {BiQUE} & {\textbf{3H-TH}} & {3E-TE-3H-TH}\\
    \clineB{1-9}{2}
    member meronym & 1 & -2.9 & .413 & .393 & \textbf{.431}& .378 & .421 & \underline{.427}  \\ 
    hypernym & 1 & -2.46 & .210 & \underline{.309} & \textbf{.310} & .289 & .304 & .303  \\ 
    has part & 1 & -1.43 & .320 & .323 & \underline{.355} & .351 & \textbf{.384} & .346  \\ 
    instance hypernym & 1 & -0.82 & .500 & .533 & \underline{.537} & \textbf{.586} & .533 & .504  \\ 
    member of domain region & 1 & -0.78 & .423 & .423 & \underline{.481} & \underline{.481} & .464 & \textbf{.500}  \\ 
    member of domain usage & 1 & -0.74 & .438 & \underline{.458} & \underline{.458} & \textbf{.479} & \underline{.458} & \underline{.458}  \\ 
    synset domain topic of & 0.99 & -0.69 & .461 & .513 & .509 & \textbf{.540} & \underline{.522} & \underline{.522}  \\ 
    \hline
    also see & 0.36 & -2.09 & .741 & .652 & .661 & .723 & .679 & .679  \\ 
    derivationally related form & 0.07 & -3.84 & .956 & .969 & .969 & .966 & .966 & .966  \\ 
    similar to & 0.07 & -1 & 1 & 1 & 1 & 1 & 1 & 1  \\ 
    verb group & 0.07 & -0.5 & .936 & .974 & .974 &.974 & .974 & .974 \\ 
    \clineB{1-9}{2}
    \end{tabular}
    }
\caption{Comparison of H@10 for WN18RR relations in high-dimensional space ($k=200$). \textbf{Bold} indicates the best score, and \underline{underline} represents the second-best score.} 
    \label{tab:high_hierarchy}
\end{table*}

\paragraph{Hierarchy.}
The hierarchy analysis aimed to examine the benefits of using hyperbolic geometry for capturing hierarchy properties. Table \ref{tab:low_hierarchy} presents the H@10 accuracy results for all relations in WN18RR, sorted by $\text{Khs}_r$, the Krackhardt hierarchy score \cite{krackhardt2014graph} and $\xi_r$, estimated graph curvature \cite{chami2020low}. A higher $\text{Khs}_r$ or lower $-\xi_r$ indicates a higher degree of hierarchy in the relations. The table confirms that the first 7 relations exhibit hierarchy, while the remaining relations do not. From the results, we observe that although Euclidean embeddings (TransE, RotatE) and hyperbolic embeddings (RotH, 3H-TH) perform similarly on non-hierarchical relations like \textit{verb group} and \textit{similar to}, hyperbolic embeddings outperform significantly on top 7 hierarchical relations. More discussion of this part refers to Appendix \ref{subsec:additional_hierarchy_results}

\paragraph{Relation Patterns.}
The relation patterns analysis aimed to assess the performance of different models on specific relation patterns.
To the best of our knowledge, no previous work in the KGE domain presents detailed results for these relation patterns, although several methods provide visualization results like \cite{sun2019rotate} or theoretical explanations for multiple patterns like \cite{cao2021dual}. We obtain the FB15K test data for \textit{symmetry}, \textit{antisymmetry}, \textit{inversion}, and \textit{composition} from \cite{sadeghi2021relational}, meanwhile, we use multiple pattern properties to classify them from the FB15K test data. The MRR results of relation patterns on FB15K in low-dimensional space (dim = 32), including \textit{symmetry}, \textit{antisymmetry}, \textit{inversion}, \textit{composition}, and \textit{multiple}, are summarized in Table~\ref{tab:relation_patterns_accuracy}. 

We can observe that the 3H-TH model outperforms on relation patterns such as \textit{symmetry}, \textit{composition}, \textit{inversion}, and \textit{multiplicity}, either achieving the best score or the second-best score. 
RotatE performs better on Symmetry and Antisymmetry because this model is simple and targeted to these two properties. Moreover, 3D rotation-based methods (3H-TH, 3E-TE-3H-TH) tend to perform better than 2D rotation-based methods (RotH, 2E-TE-2H-TH) on composition patterns in Hyperbolic space, which may indicate that 3D rotation can help the model to acquire non-commutative property on the composition pattern, although we did not classify the test data to test this. Finally, for evaluating multiple patterns, we obverse that 3H-TH can achieve the best results and combination-based methods (combine translation and rotation)(2E-TE, 3E-TE) perform better than the single-based methods (TransE, RotatE, QuatE) on the multiple patterns, which shows that combination-based methods enable model powerful representation capability of multiple patterns. (For a more comprehensive analysis of the results for the frequency distribution of various relation patterns within the datasets, please consult \ref{subsec:frequency_distribution})

\subsection{Results in high dimensions}
Table~\ref{tab:overview_result_in_high_WN18RR} displays the link prediction accuracy results for WN18RR in high-dimensional space ($k=200, 300, 500$). As anticipated, the 3H-TH model and some other composite models (2E-TE-2H-TH, 3E-TE-3H-TH) achieve new state-of-the-art (SOTA) results. However, the accuracy is comparable to that of RotH and Euclidean space methods. This indicates that Euclidean and hyperbolic embeddings perform similarly when the embedding dimension is large. 

Furthermore, Table~\ref{tab:high_hierarchy} presents the H@10 results for each relation in WN18RR using high-dimensional embeddings. In comparison to Euclidean embedding methods (TransE, RotatE), hyperbolic embedding methods (RotH, 3H-TH, 3E-TE-3H-TH) perform better on hierarchical relations such as \textit{member meronym}, \textit{hypernym}, and \textit{has part}. This indicates that hyperbolic embeddings can effectively capture and model hierarchy even in high-dimensional spaces.


\section{Conclusion}
In this study, we propose the 3H-TH model for KGE to address multiple relation patterns, including \textit{symmetry, antisymmetry, inversion, commutative composition, non-commutative composition, hierarchy,} and \textit{multiplicity.} By combining 3D rotation and translation in hyperbolic space, the model effectively represents entities and relations. Experimental results demonstrate that the 3H-TH model achieves excellent performance in low-dimensional space. Moreover, the performance difference becomes smaller in high-dimensional space, although the model still performs well.


\section*{Limitations}
\paragraph{Limited improvements in high dimensions}

While our approach 3H-TH shows substantial improvement over baseline models in a low-dimensional ($k=32$) KGE setting, we observe that as we move towards higher dimensions ($k=200, 300, 500$), our techniques tend to converge and exhibit similar results to Euclidean base models.
As an illustration, the link prediction accuracy of the 3H-TH model is similar to the Euclidean space methods, as evidenced in Table \ref{tab:overview_result_in_high_WN18RR}.
The difference in representational capacity between geometric spaces (Euclidean and hyperbolic space) becomes quite pronounced in lower dimensions. However, this gap may lessen or even disappear as the dimension is increased. 

\paragraph{Rotation in hyperbolic space}
Examining strictly from mathematical and geometric perspectives, it is correct to perform translations in hyperbolic space. However, conducting rotational operations (2D and 3D rotation) in hyperbolic space akin to those in Euclidean space lacks a certain level of rigor.

\paragraph{Time-consuming for hierarchy operations}
In Table~\ref{tab:overview_result_in_low}, all models, including those with hyperbolic operations, have space and time complexities of $O(nd+md)$ and $O(d)$ respectively, where $n$, $m$, and $d$ denote the number of entities, relations, and dimensions. Despite similar complexities, the exponential transformations and Möbius additions in hyperbolic operations notably elevate the model's computational demand. In terms of actual training time, models like RotatE that operate in Euclidean space require approximately 1/3 to 1/2 of the training time compared to the 3H-TH model.

\section*{Acknowledgements}

This study was supported by JST, the establishment of university fellowships towards the creation of science technology innovation, JPMJFS2123 (YZ).
This study was partially supported by JSPS KAKENHI 22H05106, 23H03355, and JST CREST JPMJCR21N3 (HS).
Additionally, we extend our gratitude to Junya Honda for engaging in insightful discussions and to the anonymous reviewers for their constructive feedback.
\bibliography{references}

\clearpage

\appendix
\section{Appendix}

\subsection{Hamilton's quaternions}\label{subsec:Hamilton’s_quaternions}
A quaternion $\mathrm{q}$ is composed of one real number component and three imaginary number components. It can be represented as $ \mathrm{q} = a + b \mathrm{i} + c \mathrm{j} + d \mathrm{k} $, where $a$, $b$, $c$, and $d$ are real numbers, and $\mathrm{i}$, $\mathrm{j}$, and $\mathrm{k}$ are imaginary numbers. The real part is represented by $a$, while the imaginary parts are represented by $b\mathrm{i}$, $c\mathrm{j}$, and $d\mathrm{k}$. 

Hamilton's rules govern quaternion algebra and include the following: (1). $\mathrm{i}^{2}=\mathrm{j}^{2}=\mathrm{k}^{2}=\mathrm{ijk}=-1$, (2). $\mathrm{ij}=\mathrm{k}, \mathrm{ji}=-\mathrm{k}, \mathrm{jk}=\mathrm{i}, \mathrm{kj}=-\mathrm{i}, \mathrm{ki}=\mathrm{j}, \mathrm{ik}=-\mathrm{j}$

In addition to these rules, various mathematical operations can be performed with quaternions:

\paragraph{Normalization.} 
When real elements of quaternion are numbers, $\mathrm{q}^{\triangleright} =\frac{\mathrm{q}}{|\mathrm{q}|}=\frac{a + b  \mathrm{i} + c  \mathrm{j} + d  \mathrm{k}}{\sqrt{a^{2}+b^{2}+c^{2}+d^{2}}}$. On the other hand, when the real elements of a quaternion, denoted as $\mathbf{q}_{r}$, are represented by vectors, the normalization formula needs to be modified. In this case, the quaternion normalization $\mathbf{q}^{\triangleright}_{r}$ is given by:

\[
\mathbf{q}^{\triangleright}_{r} = \frac{\mathbf{q}_{r}}{|\mathbf{q}_{r}|} = \frac{\mathbf{a} + \mathbf{b} \mathrm{i} + \mathbf{c} \mathrm{j} + \mathbf{d} \mathrm{k}}{\sqrt{\mathbf{a}^\mathsf{T} \mathbf{a} + \mathbf{b}^\mathsf{T} \mathbf{b} + \mathbf{c}^\mathsf{T} \mathbf{c} + \mathbf{d}^\mathsf{T} \mathbf{d}}}
\]

Here, $\mathbf{a}$, $\mathbf{b}$, $\mathbf{c}$, and $\mathbf{d}$ represent vector representations of the real components, and $\mathbf{a}^\mathsf{T}$, $\mathbf{b}^\mathsf{T}$, $\mathbf{c}^\mathsf{T}$, and $\mathbf{d}^\mathsf{T}$ denote the transpose of the respective vectors. The numerator consists of the vector components, and the denominator involves the Euclidean norm of the vector elements.

\paragraph{Dot product.} Given $\mathrm{q}_{1} = a_{1} + b_{1} \mathrm{i} + c_{1} \mathrm{j} + d_{1} \mathrm{k}$ and $\mathrm{q}_{2} = a_{2} + b_{2} \mathrm{i} + c_{2} \mathrm{j} + d_{2} \mathrm{k}$, we can obtain the dot product of $\mathrm{q}_{1}$ and $\mathrm{q}_{2}$:
    \[
      \mathrm{q}_{1} \cdot  \mathrm{q}_{2} 
      = a_{1} a_{2} +b_{1} b_{2} + c_{1} c_{2}+ d_{1} d_{2}.
    \]

\paragraph{Hamilton product.} The multiplication of two quaternions follows from the basic Hamilton's rule. Given $\mathrm{q}_{1}$ and $\mathrm{q}_{2}$, the multiplication is:
    \begin{align}
    \begin{split}
      \mathrm{q}_{1} \otimes \mathrm{q}_{2} & =(a_{1} a_{2}-b_{1} b_{2}-c_{1} c_{2}-d_{1} d_{2})\\
                                            & +(a_{1} b_{2}+b_{1} a_{2}+c_{1} d_{2}-d_{1} c_{2})\mathrm{i}\\
                                            & +(a_{1} c_{2}-b_{1} d_{2}+c_{1} a_{2}+d_{1} b_{2})\mathrm{j}\\
                                            & +(a_{1} d_{2}+b_{1} c_{2}-c_{1} b_{2}+d_{1} a_{2})\mathrm{k}\\  \label{eq:hamilton_product}
    \end{split}
    \end{align}
Equation (\ref{eq:hamilton_product}) presents Hamilton's product as non-commutative, which shows that 3D rotation can enable the model to perform non-commutative.

\subsection{Hyperbolic geometry}\label{subsec:Hyperbolic_Geometry}

Hyperbolic geometry, characterized by continuous negative curvature, is a non-Euclidean geometry. One way to represent hyperbolic space is through the k-dimensional Poincaré ball model with negative curvature $-\xi_r$ ($\xi_r > 0$). In this model, hyperbolic space is expressed as $\mathbb{B}_{\xi_r}^k = \{\mathbf{x} \in \mathbb{R}^k : \|\mathbf{x}\|^2 < \frac{1}{\xi_r}\}$, where $\|\cdot\|$ denotes the L2 norm. The Poincaré ball model provides a geometric framework to understand and study hyperbolic geometry.

In the Poincaré ball model, for any point $\mathbf{x} \in \mathbb{B}_{\xi_r}^k$, all possible directions of paths are contained within the tangent space $\mathcal{T}_\mathbf{x}^{\xi_r}$, which is a k-dimensional vector space. The tangent space connects Euclidean and hyperbolic space, meaning that $\mathcal{T}_\mathbf{x}^{\xi_r}\mathbb{B}_{\xi_r}^k=\mathbb{R}^k$. Since the tangent space exhibits Euclidean geometric properties, vector addition and multiplication can be performed in this space just like in Euclidean space.

\begin{figure}[t]
\centering
\begin{subfigure}[b]{1\textwidth}
        \includegraphics[width=\textwidth]{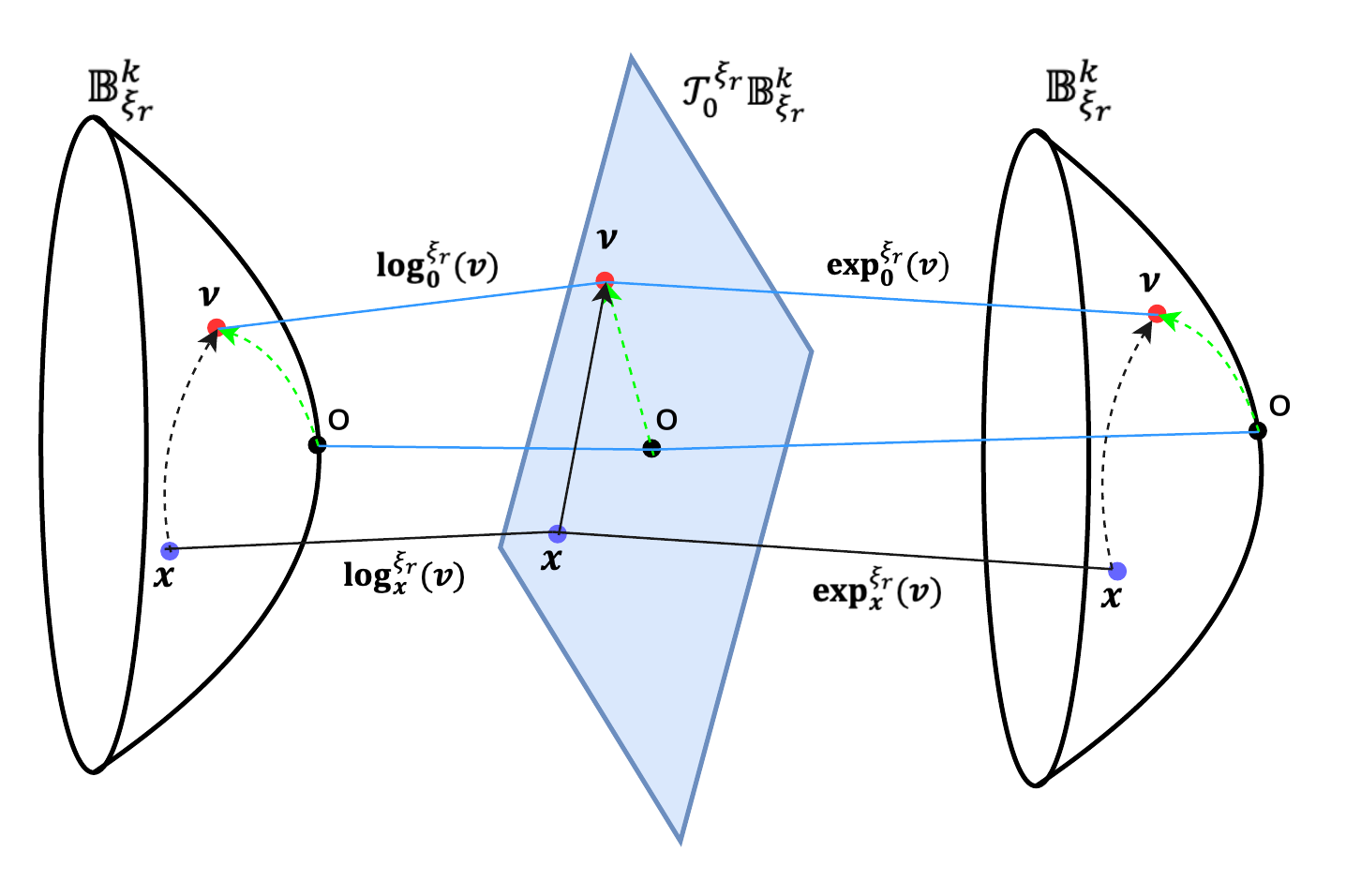}
    \end{subfigure}
\caption{
The logarithmic transformation $\mathrm{log}_\mathbf{0}^{\xi_{r}}(\mathbf{v})$ ($\mathbb{B}^{k}_{\xi_{r}}\rightarrow\mathcal{T}^{\xi_{r}}_\mathbf{0}\mathbb{B}^{k}_{\xi_{r}}$)  and the exponential transformation $\mathrm{exp}_\mathbf{0}^{\xi_{r}}(\mathbf{v})$ ($\mathcal{T}^{\xi_{r}}_\mathbf{0}\mathbb{B}^{k}_{\xi_{r}} \rightarrow \mathbb{B}^{k}_{\xi_{r}}$)
}
\label{fig:space_transformation}
\end{figure}

Moreover, the logarithmic transformation $\mathrm{log}_\mathbf{0}^{\xi_r}(\mathbf{v})$ maps a point in the Poincaré ball $\mathbb{B}_{\xi_r}^k$ to the tangent space $\mathcal{T}_\mathbf{0}^{\xi_r}\mathbb{B}_{\xi_r}^k$. Specifically, it maps a point from the origin in the direction of a vector $\mathbf{v}$. Conversely, the exponential transformation $\mathrm{exp}_\mathbf{0}^{\xi_r}(\mathbf{y})$ performs the reverse mapping. It maps a point from the tangent space $\mathcal{T}_\mathbf{0}^{\xi_r}\mathbb{B}_{\xi_r}^k$ back to the Poincaré ball, originating from the origin in the direction of a vector $\mathbf{y}$ (see Fig.~\ref{fig:space_transformation}). These transformations facilitate the conversion between the Poincaré ball and its associated tangent space, enabling geometric operations in both spaces \cite{chami2020low}.

\begin{align}
\mathrm{exp}_\mathbf{0}^{\xi_{r}}(\mathbf{v})&=\mathrm{tanh}(\sqrt{\xi_{r}}||\mathbf{v}||)\frac{\mathbf{v}}{\sqrt{\xi_{r}}||\mathbf{v}||},\label{eq:expmap}\\
\mathrm{log}_\mathbf{0}^{\xi_{r}}(\mathbf{y})&=\mathrm{tanh}^{-1}(\sqrt{\xi_{r}}||\mathbf{y}||)\frac{\mathbf{y}}{\sqrt{\xi_{r}}||\mathbf{y}||}.\label{eq:logmap}
\end{align}

We introduce the logarithmic transformation $\mathrm{log}_\mathbf{0}^{\xi_{r}}(\mathbf{v})$ ($\mathbb{B}^{k}_{\xi_{r}}\rightarrow\mathcal{T}^{\xi_{r}}_\mathbf{0}\mathbb{B}^{k}_{\xi_{r}}$) and exponential transformation $\mathrm{exp}_\mathbf{0}^{\xi_{r}}(\mathbf{y})$ ($\mathcal{T}^{\xi_{r}}_\mathbf{0}\mathbb{B}^{k}_{\xi_{r}} \rightarrow \mathbb{B}^{k}_{\xi_{r}}$) from the origin in the direction of a vector. Generally, the logarithmic transformation $\mathrm{log}_\mathbf{x}^{\xi_{r}}(\mathbf{v})$ ($\mathbb{B}^{k}_{\xi_{r}}\rightarrow\mathcal{T}^{\xi_{r}}_\mathbf{x}\mathbb{B}^{k}_{\xi_{r}}$) and exponential transformation $\mathrm{exp}_\mathbf{x}^{\xi_{r}}(\mathbf{y})$ ($\mathcal{T}^{\xi_{r}}_\mathbf{x}\mathbb{B}^{k}_{\xi_{r}} \rightarrow \mathbb{B}^{k}_{\xi_{r}}$) from $\mathbf{x}$ in the direction of a vector $\mathbf{y, v}$ respectively \cite{balazevic2019multi} are:

\[
\begin{split}
&\log _{\mathbf{x}}^{\xi_{r}}(\mathbf{y})= \\
&\frac{2}{\sqrt{\xi_{r}} \lambda_{\mathbf{x}}^{\xi_{r}}} \tanh ^{-1}\left(\sqrt{\xi_{r}}\left\|-\mathbf{x} \oplus^{\xi_{r}} \mathbf{y}\right\|\right) \frac{-\mathbf{x} \oplus^{\xi_{r}} \mathbf{y}}{\left\|-\mathbf{x} \oplus^{\xi_{r}} \mathbf{y}\right\|},\label{eq:logmap-x}
\end{split}
\]

\[
\begin{split}
&\exp _{\mathbf{x}}^{\xi_{r}}(\mathbf{v})= \\
&\mathbf{x} \oplus^{\xi_{r}}\left(\tanh \left(\sqrt{\xi_{r}} \frac{\lambda_{\mathbf{x}}^{\xi_{r}}\|\mathbf{v}\|}{2}\right) \frac{\mathbf{v}}{\sqrt{\xi_{r}}\|\mathbf{v}\|}\right).\label{eq:expmap-x}
\end{split}
\]

Besides, we apply M\"obius addition ($\mathbf{x} \oplus^{\xi_{r}} \mathbf{y}$) \cite{ganea2018hyperbolic} to replace Euclidean translation in hyperbolic space, considering that the hyperbolic space can be regarded as a roughly vectorial structure \cite{ungar2008analytic}:

\begin{align}
\begin{split}
    & \mathbf{x} \oplus^{\xi_{r}} \mathbf{y}=  \\
    & \frac{(1+2 \xi_{r} \mathbf{x}^{T} \mathbf{y}+\xi_{r}\|\mathbf{y}\|^{2}) \mathbf{x}+(1-\xi_{r}\|\mathbf{x}\|^{2}) \mathbf{y}}{1+2 \xi_{r} \mathbf{x}^{T} \mathbf{y}+{\xi_{r}}^{2}\|\mathbf{x}\|^{2}\|\mathbf{y}\|^{2}}  \label{eq:mobius_addition}
\end{split}
\end{align}

\begin{table*}[t]
\resizebox{1\textwidth}{!}{\renewcommand{\arraystretch}{1}
    \centering
    \begin{tabular}{lcccccccc}
    \clineB{1-9}{2}
     Model & MRR & H@1 & H@3 & H@10 & 1-1 (1.34\%) & 1-n (15.16\%) & n-1 (47.45\%) & n-n (36.06\%) \\
    \clineB{1-9}{2}
        3H-TH (Hyperbolic distance)      & .473 & .435 & .485 & .547 & .911 & .226 & .190 & .931 \\ 
        3H-TH (Project \& Inner product) & .356 & .342 & .362 & .380 & .703 & .057 & .029 & .900 \\
        QuatE (Inner product)            & .358 & .264 & .413 & .529 & .921 & .085 & .054 & .902 \\
        QuatE (Euclidean distance)       & .445	& .407 & .463 & .515 & .889 & .176 & .164 & .899\\
    \clineB{1-9}{2}
    \end{tabular}
    }
    \caption{The accuracy results (MRR, H@1,3,10) and complex relation MRR results (1-1, 1-n, n-1, n-n) of various scoring function methods in WN18RR.}
    \label{tab:scoring_function_results_wn18rr}
\end{table*}

\begin{table*}[t]
\resizebox{1\textwidth}{!}{\renewcommand{\arraystretch}{1}
    \centering
    \begin{tabular}{lcccccccc}
    \clineB{1-9}{2}
     Model & MRR & H@1 & H@3 & H@10 & 1-1 (1.34\%) & 1-n (15.16\%) & n-1 (47.45\%) & n-n (36.06\%) \\
    \clineB{1-9}{2}
        3H-TH (Hyperbolic distance)      & .507 & .387 & .577 & .728 & .601	& .524 & .528 & .494 \\ 
        3H-TH (Project \& Inner product) & .500	& .385 & .564 & .721 & .535	& .497 & .516 & .497 \\
        QuatE (Inner product)            & .457	& .345 & .514 & .675 & .400	& .450 & .485 & .454 \\
        QuatE (Euclidean distance)       & .484	& .360 & .556 & .715 & .578 & .504 & .515 & .480\\
    \clineB{1-9}{2}
    \end{tabular}
    }
    \caption{The accuracy results ((MRR, H@1,3,10) ) and complex relation MRR results (1-1, 1-n, n-1, n-n) of various scoring function methods in FB15K.}
    \label{tab:scoring_function_results_fb15k}
\end{table*}

\begin{table}[t]
\resizebox{1\textwidth}{!}{\renewcommand{\arraystretch}{1}
    \centering
    \begin{tabular}{lc}
    \clineB{1-2}{2}
     Relation & Num-relations(Percentage)  \\
     \clineB{1-2}{2}
        member meronym              & 253 (0.87\%) \\ 
        hypernym                    & 1251 (39.92\%)\\
        has part                    & 172 (5.49\%) \\
        instance hypernym           & 122 (3.89\%) \\
        member of domain region     & 26 (0.83\%) \\
        member of domain usage      & 24 (0.77\%) \\
        synset domain topic of      & 114 (3.64\%) \\
        also see                    & 56 (1.79\%) \\
        derivationally related form & 1074 (34.27\%) \\
        similar to                  & 3 (0.09\%) \\
        verb group                  & 39 (1.24\%) \\
    \clineB{1-2}{2}
    \end{tabular}
    }
    \caption{Frequency distribution of different relations in WN18RR.}
    \label{tab:TransE_hierarchy}
\end{table}

\begin{table*}[t]
\resizebox{0.8\textwidth}{!}{\renewcommand{\arraystretch}{1.1}
    \centering
    \begin{tabular}{llccccccc}
    \clineB{1-9}{2}
    & & \multicolumn{2}{c}{Dim = 32} & \multicolumn{5}{c}{Dim = 200} \\
     Model & MRR & H@1 & H@3 & H@10 & MRR & H@1 & H@3 & H@10 \\
    \clineB{1-9}{2}
        TransE (TE)    & .231 & .155 & .259 & .375 & .490 & .401 & .542 & .659 \\
        RotatE (2E)    & .300 & .223 & .328 & .444 & .495 & .402 & .550 & \textbf{.670} \\
        QuatE (3E)     & .380 & .302 & .421 & .544 & \underline{.516} & \underline{.435} & .560 & \underline{.661} \\
        MuRP (TH)      & .337 & .253 & .377 & .492 & .470 & .371 & .530 & .652 \\
        RotH (2H)      & .393 & .307 & .435 & .559 & .507 & .434 & .556 & .655 \\
        3H             & \underline{.401} & \underline{.314} & \underline{.440} & \underline{.562} & .511 & .432 & \underline{.563} & .631 \\
        \textbf{3H-TH} & \textbf{.409} & \textbf{.330} & \textbf{.447} & \textbf{.563} & \textbf{.520} & \textbf{.440} & \textbf{.565} & .633 \\
        
    \clineB{1-9}{2}
    \end{tabular}
    }
\caption{The link prediction accuracy results of YAGO3-10 in different dimension space ($k=32, 200$). \textbf{Bold} indicates the best score, and \underline{underline} represents the second-best score.}
    \label{tab:overview_result_YAGO3-10}
\end{table*}


\subsection{More details about Implementation} \label{subsec:implementation} 

In previous work, MuRP employed Riemannian Stochastic Gradient Descent (RSGD) \cite{bonnabel2013stochastic}, which is typically required for optimization in hyperbolic space. However, RSGD is difficult to use in real applications. Since it has been demonstrated that tangent space optimization is effective \cite{chami2019hyperbolic}, we first define all the 3H-TH parameters in the tangent space at the origin and apply conventional Euclidean methods to optimize the embeddings. Afterward, we use exponential transformation to map the parameters from Euclidean space to hyperbolic space. Therefore, all the 3H-TH model parameters $\left\{\left(\mathbf{e}_{r},  \mathbf{q}_{r}, \xi_r\right)_{r \in \mathcal{R}},\left(\mathbf{e}_{v}, b_v\right)_{v \in \mathcal{V}}\right\}$ are now Euclidean parameters that can be learned using conventional Euclidean optimization methods such as Adam or Adagrad.

Furthermore, models are trained on a single RTX8000 (48GB) GPU. For 3H-TH and related composite models, training times are approximately 1 hour for WN18RR, 4 hours for FB15K-237, and 10 hours for FB15K. We use \href{https://pytorch.org}{PyTorch} and \href{https://numpy.org}{Numpy} as the additional tools to conduct our experiment. We use \href{https://chat.openai.com/#}{ChatGPT} in our paper writing.

\subsection{Additional experiments and results } \label{subsec:additional_results}

We have included supplementary experiments in the appendix to validate our methods. 
\ref{subsec:scoring-function-discussion} focuses on comparing various scoring functions, providing additional experiments and results that demonstrate the superiority of hyperbolic-distance-based scoring functions over others. 
\ref{subsec:additional_hierarchy_results} utilizes statistical analyses of each relation to elucidate why TransE excels in specific hierarchy relations.  
Furthermore, \ref{subsec:accuracy_YAGO3-10} presents the link prediction accuracy results of YAGO3-10 in different dimension space ($k=32, 200$).
Lastly, \ref{subsec:frequency_distribution} presents the frequency distribution of various relation patterns, shedding light on the importance of each pattern.

\subsubsection{Comparison of various scoring function} \label{subsec:scoring-function-discussion}

In our 3H-TH model, we employed a distance-based scoring function (hyperbolic distance) to replace the inner-product to better utilize the advantages of the hyperbolic space, particularly its ability to better capture hierarchical properties. However, distance-based scoring function may lose the Complex Relation properties (1-1, 1-n, n-1, n-n) compared with dot product scoring function which utilized by QuatE\cite{zhang2019quaternion}. Therefore, we conduct supplementary experiments to verify which scoring function is best. 

We introduce three additional models for comparison alongside the 3H-TH model. The first model, denoted as 3H-TH (Project \& Inner product), entails transforming the head entity from hyperbolic space to Euclidean space within the 3H-TH model, utilizing the inner product as its scoring function. The second model, referred to as QuatE (Inner product), corresponds to the original QuatE model employing the dot product as its scoring function. The final model, QuatE (Euclidean distance), employs Euclidean distance as the scoring function within the QuatE model. In Table~\ref{tab:scoring_function_results_wn18rr} and \ref{tab:scoring_function_results_fb15k}, we present the overall mean reciprocal rank (MRR), overall accuracies (H@1,3,10), and MRR specifically for complex relation patterns (1-1, 1-n, n-1, n-n) in the WN18RR and FB15K datasets, respectively. The values in parentheses denote the percentages of triple instances. These experiments were conducted in a low-dimensional space (dim = 32).

Across both datasets, the 3H-TH model using hyperbolic distance consistently offers better performance than other models. Which suggesting that a hyperbolic distance-based scoring function can better utilize the strengths of hyperbolic space. Besides, when contrasting 3H-TH (Hyperbolic distance) and 3H-TH (Project \& Inner product) across both datasets, the former consistently shows better results in terms of accuracy and complex relation metrics. Finally, the performance of QuatE (Euclidean distance) surpasses QuatE (Inner product) in both datasets in low-dimensional space. This implies that, particularly in low-dimensional spaces, distance-based methods can provide a more precise measure of the differences between two vectors than inner-product based methods. In conclusion, the distance-based scoring function performs BETTER than the inner-product one in QuatE, especially in low dimensions, while they perform similarly in high dimensions. Our proposed 3H-TH uses distance in hyperbolic space and performs even better than QuatE.

\subsubsection{Explanation of TransE performs well on certain hierarchy relations} \label{subsec:additional_hierarchy_results}

Phenomena have been observed where TransE (TE) exhibits noteworthy performance on specific hierarchy relations, as exemplified in Table~\ref{tab:low_hierarchy}. Notably, the results of relations such as \textit{member meronym}, \textit{member of domain region}, and \textit{member of domain usage} indicate that TransE can achieve high accuracy, even though they cannot perform better than 3H-TH. This phenomenon can be attributed to the unbalanced distribution of individual relations within the WN18RR dataset, as demonstrated in Table~\ref{tab:TransE_hierarchy}.

As can be seen from the table, TransE methods, which perform well, such as \textit{member meronym} (8.07\%), \textit{member of domain region} (0.83\%), and \textit{member of domain uasage} (0.77\%), have a relatively low proportion in the overall test set. This can introduce an element of randomness to the results. However, in relation with a higher proportion like \textit{hypernym} (39.92\%), the performance of TransE is considerably inferior to hyperbolic methods (3H-TH, etc.). 

\subsubsection{Accuracy results on YAGO3-10 dataset}  \label{subsec:accuracy_YAGO3-10}
YAGO3-10 \cite{mahdisoltani2013yago3}, a subset of YAGO3, comprises 123,182 entities and 37 relations, predominantly describing people. We have supplemented this dataset with additional link prediction experiments. Table~\ref{tab:overview_result_YAGO3-10} displays the link prediction accuracy results for YAGO3-10 in low and high dimensional space (k = 32, 200). Our experimental results are in alignment with those obtained on datasets such as WN18RR, FB15K-237, and FB15K. Specifically, the 3H-TH model demonstrates better performance compared to all other methods in low-dimensional space (dim=32) and shows a slightly better performance in high-dimensional spaces (dim=200).

\subsubsection{Frequency distribution of various relation patterns} \label{subsec:frequency_distribution}

A pivotal aspect of our research focuses on concurrently solving various relation patterns. Consequently, it becomes imperative to delve into the statistical analysis of the frequency distribution associated with these various relation patterns within the datasets, as well as engage in a comprehensive discourse on the significance attributed to these relation patterns. In this context, we present an overview of the available data and employ specialized algorithms to calculate the frequencies of specific relation patterns embedded within the WN18RR, FB15K-237, and FB15K datasets.

\paragraph{(Anti)symmetry, Inversion, Composition}

Only (Anti)symmetry, Inversion, Composition were discovered and studied before RotatE \cite{sun2019rotate}, which provided some dataset details in their paper. In their seminal work, they elucidated that the WN18RR and FB15K237 datasets primarily encompass the \textit{symmetry}, \textit{antisymmetry}, and \textit{composition} relation patterns, whereas the FB15K dataset predominantly comprises the \textit{symmetry}, \textit{antisymmetry}, and \textit{inversion} relation patterns. Furthermore, Sadeghi et al. \cite{sadeghi2021relational} have conducted a detailed analysis of the frequency distribution of \textit{(anti)symmetry} and \textit{inversion} relation patterns within the FB15K dataset, which is presented in Table~\ref{tab:frequency_FB15K}.

From the aforementioned literature and data, it is evident that the proportion of the four relation patterns: \textit{Symmetry}, \textit{Antisymmetry}, \textit{Inversion}, and \textit{Composition}, is substantial. This underscores their research significance and value.

\begin{table}[t]
\resizebox{1\textwidth}{!}{\renewcommand{\arraystretch}{1}
    \centering
    \begin{tabular}{lccc}
    \clineB{1-4}{2}
     Triple & Symmetry & Antisymmetry & Inversion\\
     \clineB{1-4}{2}
        Train(483142) & 20333(4.2\%) & 63949(13.2\%) & 66385(13.7\%) \\ 
        Valid(50000) & 3392(6.78\%) & 25396(50.79\%) & 8798(17.60\%)\\
        Test(59071) & 3375(5.71\%) & 26020(44.05\%) & 8798(14.89\%) \\ 
    \clineB{1-4}{2}
    \end{tabular}
    }
    \caption{Frequency and proportion of (anti)symmetry and inversion in FB15K.}
    \label{tab:frequency_FB15K}
\end{table}

\begin{table}[t]
\resizebox{1\textwidth}{!}{\renewcommand{\arraystretch}{1}
    \centering
    \begin{tabular}{lcc}
    \clineB{1-3}{2}
     Dataset & Num-triples & Multiplicity \\
     \clineB{1-3}{2}
        WN18RR(Train) & 86835 & 218(0.25\%) \\ 
        WN18RR(Valid) & 3034 & 0(0.00\%) \\
        WN18RR(Test)  & 3134 & 0(0.00\%) \\ 
        FB15K-237(Train) & 272113 & 49214(18.09\%)\\
        FB15K-237(Valid) & 17535  & 160(0.91\%)\\
        FB15K-237(Test)  & 20466  & 224(1.09\%)\\
        FB15K(Train) & 483142 & 152194(31.50\%)\\
        FB15K(Valid) & 50000 & 2461(4.92\%)\\
        FB15K(Test)  & 59071 & 3341(5.66\%)\\
        
    \clineB{1-3}{2}
    \end{tabular}
    }
    \caption{Frequency and proportion of Multiplicity in WN18RR, FB15K-237, and FB15K.}
    \label{tab:frequency_multipcility}
\end{table}

\begin{table*}[t]
\resizebox{0.7\textwidth}{!}{\renewcommand{\arraystretch}{1}
    \centering
    \begin{tabular}{lccccc}
    \clineB{1-6}{2}
     Model & MRR & Std(x-y) & Var(x-y) & Se(x-y) & P-value2\\
     \clineB{1-6}{2}
        3H-TH       & .473 & - & - & - & - \\ 
        RotH(2H)    & .466 & .122 & .015 & .002 & 2.36e-03 \\
        3H          & .467 & .128 & .017 & .002 & 1.18e-05 \\
        2E-TE       & .448 & .135 & .018 & .002 & 1.14e-24 \\
        3E-TE       & .456 & .123 & .015 & .002 & 4.44e-15 \\
        2E-TE-2H-TH & .469 & .122 & .015 & .002 & 7.54e-02 \\
        3E-TE-3H-TH & .469 & .125 & .016 & .002 & 4.21e-02 \\
    
    \clineB{1-6}{2}
    \end{tabular}
    }
    \caption{Statistical significance test for 3H-TH and other baseline models in WN18RR dataset.}
    \label{tab:t-test}
\end{table*}

\begin{table*}[t]
\resizebox{\textwidth}{!}{\renewcommand{\arraystretch}{1.2}
    \centering
    \begin{tabular}{lcccc}
    \clineB{1-5}{2}
     Model & Relation embeddings & Num-params & Num-params(FB15K-237) & Num-params(FB15K)  \\
     \clineB{1-5}{2}
        TE & $\mathbf{e}_{r}$ & $ n_{e}  k + n_{r}  k$ & $14541k + 237k, (14778k)$ & $14951k + 1345k, (16296k)$ \\
        TH & $\mathbf{b}_{r}$ & $ n_{e}  k + n_{r}  k$ & $14541k + 237k, (14778k)$ & $14951k + 1345k, (16296k)$ \\ 
        2H or 2E & $\mathbf{c}_{r}$ & $ n_{e}  k + n_{r}  \frac{1}{2}k$ & $14541k + \frac{237}{2}k,(14660k)$ & $14951k + \frac{1345}{2}k, (15624k)$ \\ 
        3H or 3E & $\mathbf{q}_{r}$ & $ n_{e}  k + n_{r}  \frac{3}{4}k$ & $14541k + \frac{237*3}{4}k, (14719k)$ & $14951k + \frac{1345*3}{4}k, (15960k)$  \\ 
        2E-TE & $\mathbf{c}_{r}, \mathbf{e}_{r}$ & $ n_{e}  k + n_{r}  \frac{3}{2}k$ & $14541k + \frac{237*3}{2}k, (14897k)$ & $14951k + \frac{1345*3}{2}k, (16969k)$ \\ 
        3E-TE & $\mathbf{q}_{r}, \mathbf{e}_{r}$ & $n_{e}  k + n_{r} \frac{7}{4}k$ & $14541k + \frac{237*7}{4}k, (14956k)$ & $14951k + \frac{1345*7}{4}k, (17305k)$ \\ 
        3H-TH & $\mathbf{q}_{r}, \mathbf{b}_{r}$ & $n_{e}  k + n_{r} \frac{7}{4}k$ & $14541k + \frac{237*7}{4}k, \textbf{(14956k)}$ & $14951k + \frac{1345*7}{4}k, \textbf{(17305k)}$ \\ 
        2E-TE-2H-TH & $\mathbf{c}_{(r,E)}, \mathbf{e}_{r}, \mathbf{c}_{(r,H)}, \mathbf{b}_{r}$ & $ n_{e}  k + n_{r}  3k$ & $14541k + 237*3k, (15252k)$ & $14951k + 1345*3k, (18986k)$ \\ 
        3E-TE-3H-TH & $\mathbf{q}_{(r,E)}, \mathbf{e}_{r}, \mathbf{q}_{(r,H)}, \mathbf{b}_{r}$ & $ n_{e}  k + n_{r}  \frac{7}{2}k$ & $14541k + \frac{237*7}{2}k, (15371k)$ & $14951k + \frac{1345*7}{2}k, (19659k)$ \\ 
        
    \clineB{1-5}{2}
    \end{tabular}
    }
    \caption{The total number of parameters for several models in the FB15K-237 and FB15K datasets. $k$ denotes entity dimensions, $n_{e}, n_{r}$ denotes number of entities and relations.}
    \label{tab:additional_composite_experiments}
\end{table*}

\begin{table*}[t]
\resizebox{\textwidth}{!}{\renewcommand{\arraystretch}{1.0}
    \centering
    \begin{tabular}{lccccccc}
    \clineB{1-8}{2}
     Model & $k^{*}$(FB15K-237) & $k^{*}$(FB15K) & experiment-dim (FB15K) & MRR & H@1 & H@3 & H@10   \\
     \clineB{1-8}{2}
        TransE(TE) & 32.4 & 34 & 34 & .473 & .345 & .550 & .700\\ 
        RotatE(2E) & 32.6 & 35.4 & 36 & .474 & .354 & .540 & .706\\ 
        QuatE(3E) & 32.5 & 34.7 & 36 & .494 & .370 & .569 & .721\\ 
        MuRP(TH) & 32.4 & 34   & 34 & .490 & .361 & .561 & .721\\ 
        RotH(2H) & 32.6 & 35.4 & 36 & .505 & .380 & \underline{.585} & .729\\ 
        3H & 32.5 & 34.7 & 36 & \textbf{.520} & \textbf{.395} & \textbf{.598} & \textbf{.745}\\ 
        2E-TE & 32.1 & 32.6 & 32 & .494 & .373 & .568 & .725\\ 
        3E-TE & 32 & 32 & 32 & .496 & .376 & .572 & .725\\ 
        \textbf{3H-TH} & \textbf{32} & \textbf{32} & 32 & \underline{.506} & \underline{.383} & .581 & \underline{.731}\\
        2E-TE-2H-TH & 31.4 & 29.2 & 30 & .488 &.364 & .560 & .715\\ 
        3E-TE-3H-TH & 31.1 & 28.2 & 28 & .477 & .355 & .548 & .704\\ 
    \clineB{1-8}{2}
    \end{tabular}
    }
    \caption{The link prediction accuracy results of FB15K in different entity dimensions. \textbf{Bold} indicates the best score, and \underline{underline} represents the second-best score. $k^{*}$(FB15K-237) and $k^{*}$(FB15K) are the entity dimensions for several models under the same number of Parameters when we set that of the 3H-TH model as 32, experiment-dim denotes the dimensions that we actually use in experiments for proper experimentation.}
    \label{tab:addition_composite_results}
\end{table*}

\paragraph{Hierarchy}

Given that the hierarchy is a tree-like structure, it's challenging to provide a quantitative statistical result. Therefore, we select and compare the quantity and percentage of the top 7 more hierarchical relations in Table~\ref{tab:low_hierarchy} from the WN18RR dataset, the training set has 86,835 triples, with 62.9\% (54,603) being hierarchy relations. The test set contains 3,134 triples, 62.6\% (1,962) of which are hierarchy relations, while the validation set includes 3,034 triples, 61.6\% (1,869) of them being hierarchy relations. Based on the statistical results from WN18RR, the proportion of \textit{hierarchy} relations remains substantial. 

\paragraph{Multiplicity}

The extraction of this relation pattern is based on the properties of multiplicity, and we derived it from the dataset using the corresponding algorithm. Subsequently, we carried out statistics related to multiplicity on various datasets which has been shown in Table~\ref{tab:frequency_multipcility}.

From the statistical results in the Table~\ref{tab:frequency_multipcility}, it can be observed that on smaller datasets like WN18RR, where the number of relations is limited (number = 11), the proportion of Multiplicity relations is relatively low. However, its proportion is still significant in larger datasets like FB15K and FB15K-237, especially in the larger training sets. Thus, the Multiplicity relation patterns are also crucial and hold research significance.

\subsection{Statistical significance test} \label{subsec:t-test}

We use the WN18RR dataset for experimentation in low-dimensional space (dim = 32), the details of which can be found in Table~\ref{tab:overview_result_in_low} of the paper. And we use the MRR of each triple in 3H-TH as x, and the MRR of each triple in the other models (RotH, 3H, 2E-TE, 3E-TE, 2E-TE-2H-TH, 3E-TE-3H-TH) as y. Then, we calculated the standard deviation (Std(x-y)), variance (Var(x-y)), standard error (Se(x-y)) of the differences (x-y), and paired student's t-test (P-value2) (The test Samples are 3134, the degree of freedom is 3133, which guarantees that appropriateness of using t-test). The detailed experimental results are shown in the Table~\ref{tab:t-test}.

From the paired student's t-test results, the normal approximation (dpvalue1) is almost identical since the test sample (3134) is large. When comparing MRR and its p-value2, all the model are worse than 3H-TH. The difference are significant (p < 0.05) except for 2E-TE-2H-TH (p = 0.075). For the past model RotH (p = 0.0024 < 0.01), we can claim that RotH is significantly worse than 3H-TH. As for 2E-TE-2H-TH (p > 0.05), this model represents a novel approach that has not been proposed previously. Based on the p-value, we can assert the significant value of this model.

\subsection{Additional composite model experiments} \label{subsec:additional_composite_experiment}

The TE model has a single relation representation, denoted as $\mathbf{e}_{r}$. On the other hand, the 3E-TE-3H-TH model has four relation embeddings, namely $\mathbf{q}_{(r,E)}, \mathbf{e}_{r}, \mathbf{q}_{(r,H)}, \mathbf{b}_{r}$. Consequently, the total parameters for each model differ when we set the entity dimensions $k$ to the same value. Alternatively, we conduct additional experiments to examine the outcomes when we establish equal total parameters, encompassing both entity and relation parameters. This comparison takes into account the degrees of freedom associated with each relation type. Specifically, the translation relation $\mathbf{e}_{r}$ has $k$ parameters in each relation, the 2D rotation relation $\mathbf{c}_{r}$  has $\frac{1}{2}k$ parameters in each relation with the constraint $ \left| (\mathbf{c}_{r})_i \right| = 1 $, and the 3D rotation relation $\mathbf{q}_{r}$ has $\frac{3}{4}k$ parameters in each relation with the normalization constraint $\mathbf{q}^{\triangleright}_{r}$). For more specific information regarding the parameter counts of various models in the FB15K-237 and FB15K datasets, please refer to Table \ref{tab:additional_composite_experiments}.

We utilize the 3H-TH model as a reference and set the entity dimensions of 3H-TH to 32. The calculation of entity dimension results, denoted as $k^{*}$, for various models in the FB15K-237 and FB15K datasets, along with the link prediction accuracy results of FB15K at different entity dimensions, can be found in Table~\ref{tab:addition_composite_results}. This ensures that the overall parameters remain the same across the models. The reason for conducting experiments exclusively on FB15K, rather than FB15K-237, is that the calculation entity dimension results for FB15K-237 closely align with 32, as indicated in Table~\ref{tab:addition_composite_results}. Furthermore, WN18RR exhibits fewer relations (11) and a larger number of entities (40943) compared to FB15K-237. As a result, the calculation entity dimension results for WN18RR are also similar to 32, rendering additional experiments unnecessary. Moreover, we carefully select the appropriate dimensions for each model to ensure the proper functioning of the experiments. For instance, the dimension for 3D rotation must be a multiple of 4, while the dimension for 2D rotation is 2.

Based on the link prediction accuracy results presented in Table~\ref{tab:addition_composite_results}, it is evident that the 3H model with an entity dimension of $k=36$ surpasses all other models, including the 3H-TH model. This observation highlights the effectiveness and applicability of the 3H model in KGE tasks.

\begin{table*}[t]
\resizebox{1\textwidth}{!}{\renewcommand{\arraystretch}{1}
    \centering
    \begin{tabular}{lcc}
    \clineB{1-3}{2}
     Model & Description & MRR Accuracy \\
     \clineB{1-3}{2}
        MoCoSA\cite{he2023mocosa}       & Language Models & .696   \\ 
        SimKGC\cite{wang2022simkgc}     & Language Models & .671   \\
        LERP\cite{han2023logical}       & Additional Contextual Information (Logic Rules) & .622 \\
        C-LMKE\cite{wang2022language}   & Language Models & .598 \\
        KNN-KGE\cite{zhang2022reasoning}& Language Models & .579 \\
        HittER\cite{chen2020hitter}     & Language Models & .503 \\
        3H-TH                           & -               & .493 \\
    \clineB{1-3}{2}
    \end{tabular}
    }
    \caption{State of the art baseline models in WN18RR dataset.}
    \label{tab:SoTA_model_WN18RR}
\end{table*}

\subsection{State of the art methods in KGE} \label{subsec:SoTA_KGE}

There are several noteworthy performance methods appeared recently, and we make the following summary for WN18RR in Table~\ref{tab:SoTA_model_WN18RR}. Among them, the methods of MoCoSA\cite{he2023mocosa}, SimKGC\cite{wang2022simkgc}, C-LMKE\cite{wang2022language}, KNN-KGE\cite{zhang2022reasoning}, and HittER\cite{chen2020hitter} are mainly based on Large Language Models to complete the dataset information, thereby achieving better results. LERP\cite{han2023logical} did not use LLMs, but they used some additional contextual information (Logic Rules) beyond the dataset to complete some information missing in the entities and relations. Compared to other methods that rely on the dataset itself, for instance, TransE\cite{bordes2013translating}, RotatE\cite{sun2019rotate}, and the method 3H-TH in this paper, they only used the data and information of the KGE dataset itself, and based on certain mathematical rules and algorithms to get the final result, without using any additional information, and are not similar to LLMs' black box methods. Hence, these dataset-dependent methods continue to hold significant value for KGE research.

\subsection{Relation pattern examples} \label{subsec:relation_pattern_example}
\begin{figure}[t]
\centering
    \begin{subfigure}[b]{1\textwidth}
        \includegraphics[width=\textwidth]{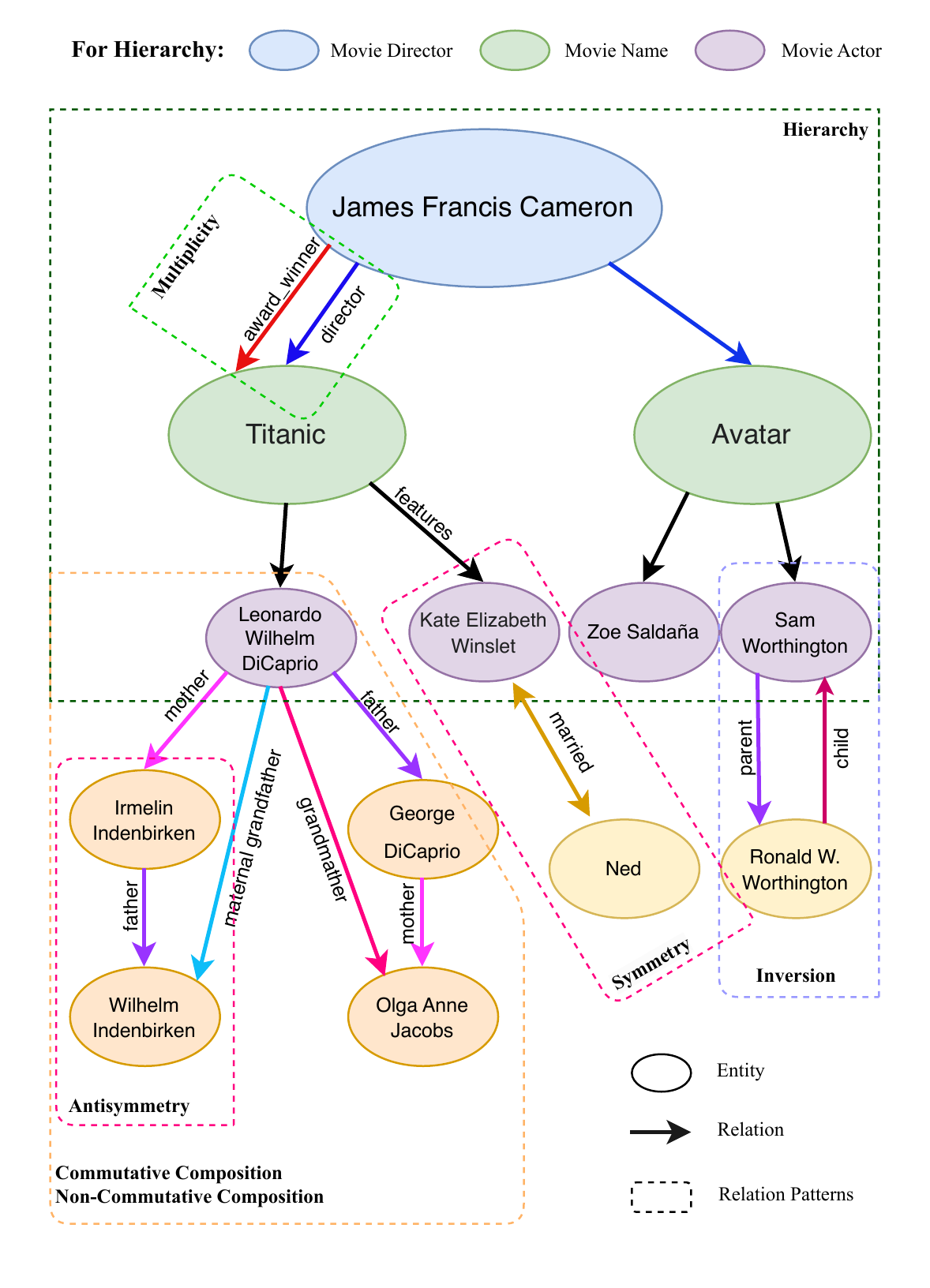}
    \end{subfigure}
    \caption{Toy examples for several relation patterns. Our approach can perform well on all these relation patterns.}
    \label{fig:toy_example_2}
\end{figure}

In knowledge graphs (KGs), various relation patterns can be observed, including symmetry, antisymmetry, inversion, composition (both commutative and non-commutative), hierarchy, and multiplicity. These patterns are illustrated in Fig.~\ref{fig:toy_example_2}.

Some relations exhibit symmetry, meaning that if a relation holds between entity $x$ and $y$ ($\left(r_{1}(x, y) \Rightarrow r_{1}(y, x)\right)$)(e.g., \textit{is married to}), it also holds in the reverse direction (i.e., between $y$ and $x$). On the other hand, some relations are antisymmetric ($\left(r_{1}(x, y) \Rightarrow \neg r_{1}(y, x)\right)$), where if a relation holds between $x$ and $y$ (e.g., \textit{is father of}), it does not hold in the reverse direction (i.e., between $y$ and $x$).

Inversion ($\left(r_{1}(x, y) \Leftrightarrow r_{2}(y, x)\right)$) of relations is also possible, where one relation can be transformed into another by reversing the direction of the relation (e.g., \textit{is child of} and \textit{is parent of}).

Composition ($\left(r_{1}(x, y) \cap r_{2}(y, z) \Rightarrow r_{3}(x, z)\right)$) of relations is another important pattern, where the combination of two or more relations leads to the inference of a new relation. This composition can be commutative (order-independent) or non-commutative (order-dependent). Non-commutative composition ($\left(r_{1}(x, y) \cap r_{2}(y, z) \neq \right(r_{2}(x, y) \cap r_{1}(y, z)$) is necessary when the order of relations matters, such as in the example of the mother of A's father (B) being C and the father of A's mother (D) being E. In a commutative composition, C and E would be equal, but in a non-commutative composition, they are not.

Hierarchical relations exist in KGs, where different entities have different levels or hierarchies. This hierarchical structure is depicted in the tree-like structure shown in Fig.~\ref{fig:toy_example_2}.

Finally, multiplicity refers to the existence of different relations between the same entities. For example, an entity can have multiple relations such as \textit{award-winner} and \textit{director} associated with it.

These various relation patterns capture the complexity and diversity of knowledge in KGs, highlighting the challenges and opportunities in modeling and reasoning over such data.

\subsection{Hyperparameter}
All the hyperparameter settings have been shown in Table~\ref{tab:hyperparameter}.
\begin{table*}[t]
\centering
\small
\begin{tabular}{l|clcccc}
\clineB{1-7}{2}
{Dataset} &  embedding dimension & model & {learning rate} & {optimizer} & {batch size} & {negative samples} \\
\clineB{1-7}{2}
\multirow{33}{*}{{WN18RR}} & \multirow{11}{*}{{32}} 
& TransE(TE) & 0.001 & Adam & 500 & 50 \\
&  & RotatE(2E) & 0.1 & Adagrad & 500 & 50 \\
&  & QuatE(3E) & 0.2 & Adagrad & 500 & 50\\
&  & MuRP(TH)  & 0.0005 & Adam & 500 & 100\\
&  & RotH(2H) & 0.0005 & Adam & 500 & 50 \\
&  & 3H    & 0.001 & Adam & 500 & 100 \\
&  & 2E-TE & 0.1 & Adagrad & 500 & 50 \\
&  & 3E-TE & 0.2 & Adagrad & 500 & 100 \\
&  & 2E-TE-2H-TH & 0.001 & Adam & 500 & 100 \\
&  & \textbf{3H-TH} & 0.001 & Adam & 500 & 100 \\
&  & 3E-TE-3H-TH & 0.001 & Adam & 500 & 100 \\
\cline{2-7}
& \multirow{11}{*}{{200}} 
& TransE(TE) & 0.001 & Adam & 500 & 100 \\
&  & RotatE(2E) & 0.1 & Adagrad & 500 & 100 \\
&  & QuatE(3E) & 0.2 & Adagrad & 500 & 100\\
&  & MuRP(TH) & 0.001 & Adam & 500 & 100 \\
&  & RotH(2H) & 0.001 & Adam & 500 & 50 \\
&  & 3H    & 0.001 & Adam & 500 & 100 \\
&  & 2E-TE & 0.1 & Adagrad & 500 & 50 \\
&  & 3E-TE &  0.2 & Adagrad & 500 & 100 \\
&  & 2E-TE-2H-TH & 0.001 & Adam & 500 & 100 \\
&  & \textbf{3H-TH} & 0.001 & Adam & 500 & 100 \\
&  & 3E-TE-3H-TH & 0.001 & Adam & 500 & 100 \\
\cline{2-7}
& \multirow{11}{*}{{300, 500}} 
& TransE(TE) & 0.001 & Adam & 500 & 100 \\
&  & RotatE(2E) & 0.1 & Adagrad & 500 & 100 \\
&  & QuatE(3E) & 0.2 & Adagrad & 500 & 100\\
&  & MuRP(TH) & 0.001 & Adam & 500 & 100 \\
&  & RotH(2H) & 0.001 & Adam & 500 & 50 \\
&  & 3H    & 0.001 & Adam & 500 & 100 \\
&  & 2E-TE & 0.1 & Adagrad & 500 & 50 \\
&  & 3E-TE & 0.2 & Adagrad & 500 & 100 \\
&  & 2E-TE-2H-TH & 0.001 & Adam & 500 & 100 \\
&  & \textbf{3H-TH} & 0.001 & Adam & 500 & 100 \\
&  & 3E-TE-3H-TH & 0.001 & Adam & 500 & 100 \\
\clineB{1-7}{2}
\multirow{11}{*}{{FB15k-237}} & \multirow{11}{*}{{32}} 
& TransE(TE) & 0.05 & Adam & 1000 & 50 \\
&  & RotatE(2E) & 0.05 & Adagrad & 1000 & 50 \\
&  & QuatE(3E)& 0.05 & Adagrad & 1000 & 50\\
&  & MuRP(TH) & 0.05 & Adagrad & 1000 & 50 \\
&  & RotH(2H) & 0.1 & Adagrad & 1000 & 50 \\
&  & 3H & 0.05 & Adagrad & 1000 & 50 \\
&  & 2E-TE & 0.05 & Adagrad & 1000 & 50 \\
&  & 3E-TE & 0.05 & Adagrad & 1000 & 50 \\
&  & 2E-TE-2H-TH & 0.05 & Adagrad & 1000 & 50 \\
&  & \textbf{3H-TH}& 0.05 & Adagrad & 1000 & 50 \\
&  & 3E-TE-3H-TH & 0.05 & Adagrad & 1000 & 50 \\
\clineB{1-7}{2}
\multirow{11}{*}{{FB15K}} & \multirow{11}{*}{{32}} 
& TransE(TE) & 0.05 & Adagrad & 1000 & 200 \\
&  & RotatE(2E) & 0.4 & Adagrad & 1000 & 200 \\
&  & QuatE(3E) & 0.2 & Adagrad & 1000 & 200\\
&  & MuRP(TH) & 0.1 & Adagrad & 1000 & 200 \\
&  & RotH(2H) & 0.1 & Adagrad & 1000 & 200 \\
&  & 3H & 0.2 & Adagrad & 1000 & 200 \\
&  & 2E-TE & 0.4 & Adagrad & 1000 & 200 \\
&  & 3E-TE & 0.2 & Adagrad & 1000 & 200 \\
&  & 2E-TE-2H-TH & 0.2 & Adagrad & 1000 & 200 \\
&  & \textbf{3H-TH} & 0.2 & Adagrad & 1000 & 200 \\
&  & 3E-TE-3H-TH & 0.2 & Adagrad & 1000 & 200 \\
\clineB{1-7}{2}
\end{tabular}
\caption{Best hyperparameters in low- and high-dimensional settings for our approach and several composite models.} 
\label{tab:hyperparameter}
\end{table*}

\end{document}